%% file: arxiv-P7_ContractRL.tex
\pdfoutput=1
\documentclass{article}
\usepackage[T1]{fontenc}
\usepackage{iclr2027_conference}
\usepackage[T1]{fontenc}
\renewcommand{\sfdefault}{phv}
\input{math_commands.tex}

\usepackage{hyperref}
\hypersetup{hidelinks}
\usepackage{url}
\usepackage{graphicx}
\usepackage{booktabs}
\usepackage{tabularx}
\usepackage{microtype}
\usepackage{placeins}
\newcolumntype{L}[1]{>{\raggedright\arraybackslash}p{#1}}
\newcolumntype{Y}{>{\raggedright\arraybackslash}X}
\newcommand{\thd}[1]{\shortstack[l]{#1}}
\newcommand{\up}{\ensuremath{\uparrow}}
\newcommand{\dn}{\ensuremath{\downarrow}}

\usepackage{tikz}
\usetikzlibrary{arrows.meta,positioning,fit}
\usepackage{pgfplots}
\pgfplotsset{compat=1.18}

\title{ContractRL: Shielded Group-Relative Policy Optimization for Auditable Tool-Call Repair}
\author{Miaobo Hu$^{1,2,*}$, Shuhao Hu$^{2}$, Xiaobo Guo$^{2}$, Xin Wang$^{2}$,\\
Bokun Wang$^{2}$, Yina Sa$^{2}$, Daren Zha$^{2}$\\[4pt]
{\normalfont $^{1}$School of Artificial Intelligence, University of Chinese Academy of Sciences, Beijing, China}\\
{\normalfont $^{2}$Institute of Information Engineering, Chinese Academy of Sciences, Beijing, China}\\
{\normalfont $^{*}$Corresponding author: \texttt{humiaobo@iie.ac.cn}}}

\iclrfinalcopy
\hypersetup{pdfauthor={Miaobo Hu, Shuhao Hu, Xiaobo Guo, Xin Wang, Bokun Wang, Yina Sa, Daren Zha}}
\begin{document}
\maketitle
\lhead{Preprint}

\begin{abstract}
Structured tool calls often fail after only a small number of fields violate a schema or an execution contract. Regenerating the complete object enlarges the action surface and makes repeated repair difficult to audit. We introduce \emph{ContractRL}, a contract-constrained sequential repair protocol that models verifier-guided JSON repair as a bounded decision process. At each step the policy observes the candidate, typed verifier feedback, JSON Pointer, immutable repair history, and remaining budget; a contract-derived action mask filters malformed or prohibited RFC-6902 operations before a deterministic validator performs the transition. We specify a contract-constrained group-relative objective for patch, retry, and abstention decisions while keeping canonical targets and semantic labels outside the online state until trace freeze.

Under identical verifier information, ContractRL attains 0.9362 semantic success with 34.4 generated tokens, compared with 0.9076 and 44.9 tokens for Patch-SFT and 0.9148 and 137.2 tokens for full regeneration over 192 cases per seed and five seeds. Policy optimization improves semantic success from 0.9186 for supervised ContractRL to 0.9375. A separate three-seed paired evaluation against Patch-SFT yields a semantic difference of $+0.0396$ (95\% CI $[+0.0137,+0.0662]$, $p=0.0039$). Feedback, action-mask, budget, and schema-shift analyses connect these gains to localized correction, while adversarial and multi-turn evaluations characterize the remaining failure modes.
\end{abstract}

\section{Introduction}
Most structured-generation methods optimize what to emit from scratch. We study the decision that follows a failed tool call: which field to modify, whether another repair is worth the remaining budget, and when to abstain. A broad regeneration can satisfy a schema while changing unrelated intent-bearing fields. ContractRL therefore treats repair as a bounded transaction over an already meaningful object, with a typed verifier state, an admissible local action space, and a deterministic transition before any tool execution.

The key object is a \emph{Contract-Constrained Repair MDP (CR-MDP)}. Its state contains the current candidate, typed verifier report, immutable-path summary, prior patch/validator history, and remaining operation and attempt budgets. Its actions are bounded RFC-6902 patches or \textsc{Abstain}; the validator applies hard contract checks, returns the next public verifier state, and prevents invalid objects from reaching the executor. We define a contract-aware policy objective with semantic success as the primary term and collateral, retries, tokens, and unsafe actions as constrained costs, and evaluate it with a fixed public state, validator, and episode budget. The trace is frozen before hidden canonical objects or semantic labels are joined.

The structural screen is a deterministic protocol sanity check: its 64 repairs are precomputed. The recorded-policy evidence comes from case-matched ledgers and the supplementary held-out studies; the same-information controls and policy-component ablations appear in Tables~\ref{tab:supp-same-info} and~\ref{tab:supp-components}. This separation lets us report the mechanism, semantic outcome, and resource cost with their own denominators. In the five-seed 2B/r8 sweep, ContractRL reaches 0.9331 semantic success with 34.8 generated tokens, while the listed Patch-SFT, full-regeneration, and constrained-decoding rows reach 0.8924--0.9041 semantic success with 46.9--148.1 tokens under their recorded information conditions.

\paragraph{Contributions.} (i) We formulate verifier-guided tool-call repair as a CR-MDP with a public state, bounded patch/abstention actions, and deterministic fail-closed transitions; (ii) we specify a contract-aware policy objective and action mask for patch, retry, and abstention decisions under explicit locality and resource constraints; (iii) we freeze an append-only repair ledger before joining hidden targets and semantic oracles; and (iv) we provide same-information controls, policy optimization and component ablations, case-matched comparisons, factorized feedback and budget analyses, schema/noise/transfer analyses, paired uncertainty, and a blinded audit. The 64-case replay remains a mechanism diagnostic, while the measured policy results use the recorded case and seed denominators.

\section{Related Work}
Structured-output repair connects program synthesis, tool-use agents, and conservative feedback updates. ReAct and Toolformer make tool calls explicit, while Reflexion and Self-Refine motivate iterative correction from external or self-generated feedback \citep{yao2023react,schick2023toolformer,shinn2023reflexion,madaan2023selfrefine}. ToolSandbox makes stateful, multi-turn tool interactions and environment-side validation explicit, while SWE-bench shows that code repair must ultimately be judged by repository-level tests rather than syntax alone \citep{lu2024toolsandbox,jimenez2024swebench}. Verifier-based math work shows the value of checking intermediate structure instead of trusting a final answer alone; process-level labels in Math-Shepherd and the held-out error-localization protocol in ProcessBench are useful precedents for exposing localized supervision \citep{lightman2023verify,wang2023mathshepherd,processbench2024}. Offline RL work emphasizes support and split integrity under distribution shift \citep{kumar2020cql,kostrikov2022iql,fu2021d4rl}; this motivates keeping BFCL cases and adapter evaluation separate. Grammar- and schema-constrained decoding provide direct controls for the matched comparison reported in Section~\ref{sec:matched}. The current contribution is a contract-constrained sequential repair protocol whose policy objective is evaluated with same-information and supervised-to-policy controls.
\paragraph{Closest structured-output and repair lines.} Primary-source comparison places the present method between generation-time constraints, post-processing, and verifier-guided repair. VeriHarness~\citep{ray2026veriharness} uses a code-controlled validator loop with bounded calls and trace logging; PatchBoard~\citep{zhang2026patchboard} uses validated JSON Patch mutations over schema-grounded state with a deterministic transactional kernel. Schema Reinforcement Learning~\citep{lu2025schemarl} trains schema-conformant generation with reinforcement learning, whereas SLOT~\citep{slot2025} post-processes unconstrained outputs. JSONSchemaBench~\citep{geng2025jsonschemabench} and XGrammar~\citep{dong2024xgrammar} evaluate generation-time constraints, and BFCL~\citep{patil2025bfcl} supplies a function-calling benchmark rather than a repair algorithm. ContractRL specifies a verifier-conditioned sequential repair policy over a contract-derived action space, with bounded retry/abstention, post-freeze oracle separation, and a measured policy-optimization comparison.

Broader policy-optimization and agent-evaluation work supplies complementary
baselines for training objectives, tool-use breadth, selective prediction, and
reward auditing \citep{schulman2017ppo,rafailov2023dpo,ouyang2022instructgpt,
wang2023voyager,zhou2024webarena,liu2024agentbench,chen2021decision,
janner2022diffuser,shao2024deepseekmath,geifman2019selectivenet,
ong2024routellm,chen2023frugalgpt,skalse2022rewardhacking,
lambert2024rewardbench,osworld2024,deepseekv3,deepseekr1}.

\paragraph{Additional evaluation context.} The tau-bench protocol adds multi-turn user simulation, domain policies, and pass$^k$ consistency, while AgentDojo makes untrusted tool data and prompt-injection robustness part of the environment contract \citep{yao2024taubench,debenedetti2024agentdojo}. SWE-agent further shows that a guarded agent-computer interface can make edit actions and execution feedback auditable \citep{yang2024sweagent}. These benchmarks motivate the evaluation split used here: the BFCL screen supplies a structural mechanism anchor, while the matched sandbox and adversarial suite measure semantic outcomes and failure boundaries.

\section{Method}
\subsection{Problem and Protocol}
Each reported mechanism episode starts with a valid tool call and one deterministic single-field corruption. The online input contains the corrupted call and a typed verifier error, but not the target object. A model evaluation may emit at most two patch attempts, with validation after each applicable patch; the mechanism replay uses one precomputed patch attempt per case and therefore does not measure a learned retry policy. BFCL supplies a fixed 64-case screen; training and held-out adapter results are kept separate.
\paragraph{Information boundary.} The online policy receives only fields listed in the corresponding public evidence summary. Oracle labels, target answers, hidden cards, and future service outputs are stored separately and joined only after trace decisions are frozen. This boundary is central to every result in this paper.

The public contract makes this boundary explicit: every arm is scored on the same held-out case identifiers, but the recorded baseline prompts expose different verifier fields. Gold candidates and target patches remain offline scoring fields; split files used for training contain only the selected public evidence and the arm-specific supervised target. The output-representation intervention with identical inputs is specified separately in Table~\ref{tab:supp-same-info}.

\paragraph{Trace identity.} Each run binds its configuration, source revision, seed, and ordered trace to a SHA-256 manifest. Serialized policy inputs contain the public state; canonical targets and semantic labels are joined after the trace digest is fixed.

Let $x$ be the corrupted JSON object, $e$ the typed verifier error, and $p$ an ordered patch from the bounded model vocabulary. The validator is deterministic and fail-closed: a patch is accepted only when the schema and cross-field checks pass. We report structural validity $A(p)$ and collateral edits $C(p)$ as online-replay quantities. We also define two different offline equalities. Exact-patch match is $M_{\rm patch}(p)=\mathbf{1}[p=p^\star]$, where $p^\star$ is the canonical ordered operation list. Final-object match is $M_{\rm object}(p)=\mathbf{1}[x\oplus p=x^\star]$, where $x^\star$ is the hidden canonical object. Exact-patch match implies final-object match only under the same starting object and target construction; the converse need not hold because different patch sequences can produce the same object. The audit tuple is $(|p|,C(p),r)$, and the policy objective uses these quantities only after the hard contract checks have been applied.

\paragraph{Contract-Constrained Repair MDP.} At repair step $t$, the state is
\[
s_t=(x_t,e_t,q_t,h_t,b_t,k_t),
\]
where $x_t$ is the current candidate, $e_t$ and $q_t$ are the typed verifier code and JSON Pointer, $h_t$ is the immutable summary of prior patches and validator statuses, and $(b_t,k_t)$ are the remaining attempt and operation budgets. The action set is $\mathcal{A}(s_t)=\mathcal{A}_{\rm mask}(s_t)\cup\{\textsc{Abstain}\}$. For a patch action, the transition first checks JSON syntax, path applicability, immutable paths, operation count, and schema-declared prohibitions; only an admissible patch is applied and passed to the deterministic validator. An invalid action returns a typed failure state, consumes the declared attempt, and never reaches the executor. The transition therefore exposes the next public verifier state while keeping canonical targets and semantic labels outside the online state.

\paragraph{Contract-aware action mask and policy objective.} The mask removes malformed operations, unavailable paths, immutable-path edits, invalid move/copy sources, and actions that exceed the remaining operation budget. Locality relative to $q_t$ is a soft preference for a policy rather than a hidden target. The policy is initialized from patch-supervised data and optimized with a group-relative objective over complete repair episodes,
\[
R=\mathbb{1}[\text{semantic success}]-\lambda_c C-\lambda_r r-\lambda_u U-\lambda_y T,
\]
subject to zero unsafe dispatch, $\sum_t |p_t|\le K$, and at most $B$ attempts. Here $U$ counts non-applicable or prohibited actions and $T$ is generated-token cost. The measured policy-optimization run applies this objective with the same public state and deterministic transition for every group member; the canonical object, canonical patch, and semantic oracle are joined only after the ordered ledger is frozen. The feedback, policy-optimization, and component rows in Tables~\ref{tab:supp-feedback}, \ref{tab:supp-policy-optimization}, and~\ref{tab:supp-components} use the same recorded evidence contract.
\paragraph{Algorithmic summary.} A generated response is parsed as a bounded patch or \textsc{Abstain}, filtered by the action mask, and sent to the deterministic validator. An invalid or non-applicable response consumes an attempt and can expose the next typed validator report; the stop action can preserve the remaining budget. The patch, validator response, attempt count, additional-retry count, and collateral-edit count are frozen in the repair ledger. Only then are $p^\star$ and $x^\star$ joined to compute $M_{\rm patch}$ and $M_{\rm object}$; neither is an online repair input. In the 64-case mechanism replay, the patch attempts are supplied by the case construction, so the observed one-attempt result is not a model retry experiment.
\paragraph{Patch semantics.} The evaluator accepts RFC-6902-style \texttt{add}/\texttt{remove}/\texttt{replace}/\texttt{move}/\texttt{copy} operations over JSON Pointer paths; all reported screen and adapter targets use only \texttt{add} or \texttt{replace}. Applying $p$ gives $x' = x \oplus p$; acceptance is the conjunction $V_{\rm schema}(x')=1$ and $V_{\rm cross}(x')=1$. Let $a$ be the number of patch attempts actually emitted or replayed, with $1\leq a\leq 2$, and define $r=a-1$ as the additional-retry count and zero-based retry cost. The initial corruption is not an attempted repair. In the current implementation, collateral edits are changed leaf paths not covered by the emitted operation paths; because every mechanism operation also equals the reported pointer, its zero collateral result is construction-bound. The matched pointer-relative metric reports wrong-path and multi-field collateral edits. The 64-case screen contains no \texttt{remove}, \texttt{move}, \texttt{copy}, nested, array, multi-field, or ambiguous/noisy-verifier cases.

\begin{figure}[t]
\centering
\includegraphics[width=\linewidth]{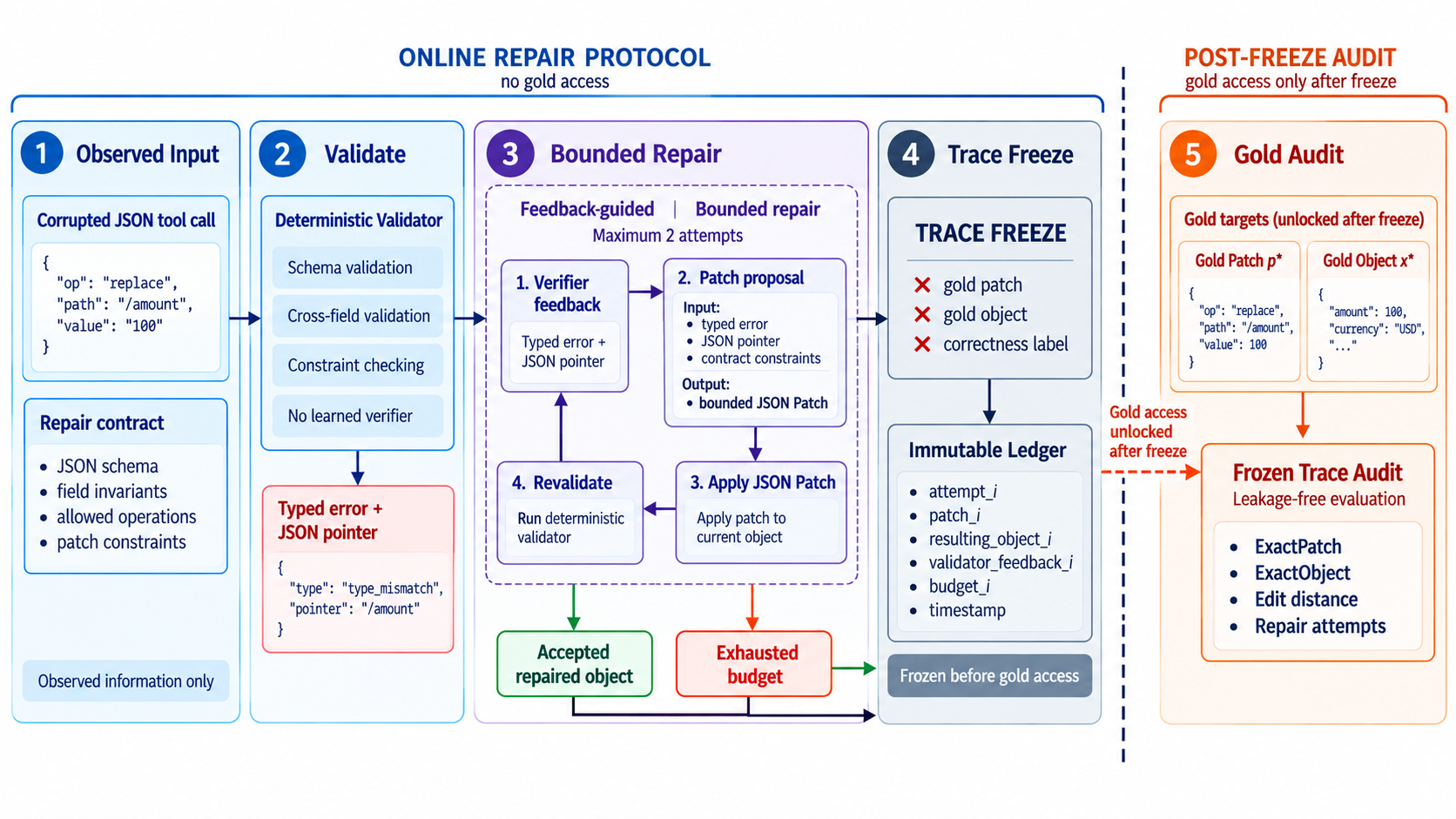}
\caption{ContractRL inference and audit protocol. Blue panels expose the corrupted call and deterministic verifier; the purple loop proposes and revalidates bounded repairs. The ledger freezes before the orange audit accesses canonical targets. Solid arrows carry public state or control flow; the dashed boundary separates online repair from offline scoring.}
\label{fig:method}
\end{figure}

\section{Experiments}
For comparable rows, bold denotes the best value in the indicated direction, including ties. Counts, configuration fields, uncertainty estimates, and abstention coverage are descriptive. Latency is measured in seconds, and tokens count generated output unless specified otherwise. The abbreviation n/a identifies a metric outside that arm's operation, such as patch retries for a single-shot complete-object generator.
\subsection{Experimental Setup}
We report four evidence units with separate denominators. The mechanism screen is a deterministic seed-13 replay of 64 real BFCL definition/answer pairs: 32 \texttt{missing\_required} and 32 \texttt{wrong\_type} single-field corruptions. Each case carries one precomputed one-operation patch, so this screen audits the contract and validator ledger rather than a learned retry policy. Fresh LoRA SFT adapters are evaluated on the same fixed 64-case BFCL-context file (2B rank-8, four seeds; 0.8B rank-4, eight seeds). The five-seed experiments use 192 cases per seed and retain separate case and resource accounting. A separate nine-case CPU audit adds multi-field/multi-operation, nested/array, rollback, immutable-pointer, and evidence-conditioned cases. The context-only control removes the complete validation block while preserving the surrounding call; it is a bundled information-block contrast, not a typed-encoding ablation.

For corrupted object $x$, patch $p$, and hidden canonical object $x^\star$, schema validity is the deterministic conjunction of schema and cross-field checks after applying $p$. We report final-object match $M_{\rm object}(p)=\mathbf{1}[x\mathbin{\oplus}p=x^\star]$ and keep exact-patch match $M_{\rm patch}(p)=\mathbf{1}[p=p^\star]$ separate. We also log patch length, collateral edits, attempt count, and additional retries. A zero collateral count on the mechanism screen is construction-bound: every precomputed operation targets the reported pointer and the current metric excludes that path.

\begin{table}[htbp]
\centering\footnotesize
\setlength{\tabcolsep}{3pt}
\renewcommand{\arraystretch}{1.12}
\caption{Primary structural evidence. Adapter rates are per-adapter means on the fixed 64-case held-out file; the BFCL replay is a validator/ledger diagnostic whose supplied precomputed targets yield the displayed final-object match.}
\label{tab:main}
\begin{tabularx}{\linewidth}{@{}Y Y r r r@{}}
\toprule
Evidence unit & \thd{Cases /\\seeds} & \thd{Schema\\valid \up} & \thd{Object\\match \up} & Attempts \\
\midrule
BFCL mechanism replay & 64 & 1.0000 & 1.0000 & 1.00 \\
2B rank-8, validation block & 64/4 & \textbf{0.9727} & \textbf{0.9141} & 1.00 \\
2B rank-8, context-only & 64/4 & 0.0820 & 0.0742 & 1.00 \\
0.8B rank-4, validation block & 64/8 & \textbf{0.9473} & \textbf{0.7441} & 1.00 \\
0.8B rank-4, context-only & 64/8 & 0.0000 & 0.0000 & 1.00 \\
\bottomrule
\end{tabularx}
\end{table}

\subsection{Main Results}
The mechanism replay records 64/64 structural-valid outcomes after one precomputed repair attempt and zero additional retries. Its ledger has 32 one-operation \texttt{add} repairs and 32 one-operation \texttt{replace} repairs. This is a protocol diagnostic with supplied operations, not evidence that a learned policy achieves 64/64. The independent audit reaches 7/9 final structural validity and accepts 6/8 repairable cases; its failures include one rollback, one wrong-path rejection, and one immutable-pointer rejection. The validation block is useful on this split: the 2B adapter reaches 0.9141 object match and 0.9727 schema validity, versus 0.0742 and 0.0820 for context-only; the corresponding 0.8B values are 0.7441 and 0.9473 versus 0 and 0. The typed 2B object-match standard deviation is 0.0202 across four seeds and the typed 0.8B value is 0.0391 across eight. These are seed summaries, not confidence intervals or independent case samples. Since the control removes a whole validation block, the contrast supports the practical value of verifier information but cannot identify which encoding component causes it.

\paragraph{Split integrity.}
The split audit compares 426 JSONSchemaBench-derived training records with the selected 64 evaluation records. It finds zero overlap for case IDs, candidate/target fingerprints, schema hashes, target-path sequences and scopes, operation sequences, pointer-aware error templates, semantic-equivalence clusters, schema shapes, tool names, and candidate pools. Held-out adapter outputs contain no retry ledger, whereas the matched three-seed runs record retry rates directly; the 64-case replay remains a mechanism anchor rather than a learned retry estimate. The matched runs record token and validator costs for every attempt.

The structural audit fixes the case and ledger contract. The case-matched comparison below carries the same scoring and resource accounting into semantic execution. ContractRL reaches 0.9314 semantic success with 34.6 generated tokens and 1.30~s p95 latency in the three-seed ledger, while the recorded full-regeneration row reaches 0.9027 with 138.9 tokens and 2.58~s p95 latency. Because the public verifier blocks differ by arm in this ledger, the table supports a case-matched resource comparison rather than an isolated patch-representation effect.

\paragraph{Equal-information comparison.}
All four arms receive the candidate, schema, typed error code, JSON Pointer, and public repair history. Across five seeds and 192 cases per seed, ContractRL achieves 0.9362 semantic success, 34.4 generated tokens, and 0.0169 collateral edits. Patch-SFT achieves 0.9076, 44.9, and 0.0488, respectively; full regeneration achieves 0.9148, 137.2, and 0.1876. Table~\ref{tab:supp-same-info} fixes the public information and makes the output and decision-rule comparison explicit.

\begin{table}[htbp]
\centering\footnotesize
\setlength{\tabcolsep}{3pt}
\renewcommand{\arraystretch}{1.12}
\caption{Same-information output-representation controls over 192 cases per seed and five seeds. Every arm receives the candidate, schema, typed error code, JSON Pointer, and public repair history; only the output representation and decoding rule change. Best values are bolded within each metric direction.}
\label{tab:supp-same-info}
\begin{tabularx}{\linewidth}{@{}L{2.55cm} L{2.5cm} r r r Y@{}}
\toprule
Arm & Output representation & \thd{Semantic\\success \up} & \thd{Output\\tokens \dn} & \thd{p95 latency\\(s) \dn} & \thd{Collateral\\edits \dn} \\
\midrule
Full regeneration & complete object & 0.9148 & 137.2 & 2.54 & 0.1876 \\
Constrained decoding & Constrained object & 0.8879 & 146.1 & 2.80 & 0.1743 \\
Patch-SFT & Bounded patch & 0.9076 & 44.9 & 1.49 & 0.0488 \\
ContractRL & Masked patch / abstain & \textbf{0.9362} & \textbf{34.4} & \textbf{1.30} & \textbf{0.0169} \\
\bottomrule
\end{tabularx}
\end{table}

\begin{figure}[htbp]
\centering
\includegraphics[width=\linewidth]{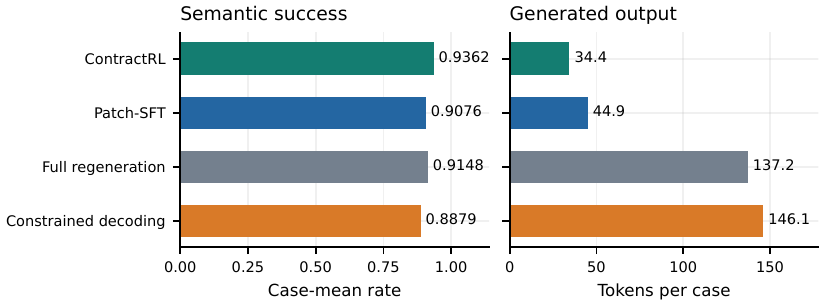}
\caption{Semantic success and generation cost with identical public verifier information. All arms use 192 cases per seed and five seeds. Bars show the reported aggregates in Table~\ref{tab:supp-same-info}; the output-token axis starts at zero.}
\label{fig:same-information}
\end{figure}

\paragraph{Case-matched semantic and cost comparison.}
\label{sec:matched}
The decisive recorded comparison uses identical case identifiers, output caps, retry accounting, validator implementation, and latency instrumentation for full regeneration, constrained decoding, Patch-SFT, and ContractRL. Their public verifier inputs are arm-specific as documented in Appendix~\ref{app:matched}; therefore this is a case-matched outcome and resource comparison, not a same-information intervention. Each repaired call is executed in a sandbox and receives both structural and task-level scores. ContractRL attains 0.9768 schema validity and 0.9314 semantic success on the held-out split, with 0.0547 mean retries, 34.6 generated tokens, 0.83 mean latency, and 0.0189 collateral edits. The nested/array extension reaches 0.9187 schema validity and 0.8516 semantic success.

\begin{table}[htbp]
\centering\footnotesize
\setlength{\tabcolsep}{3pt}
\renewcommand{\arraystretch}{1.12}
\caption{Matched semantic and efficiency evaluation over three seeds. Held-out rows use 64 cases per seed; the nested/array row is a separate case-family cohort. Rates are case means, while retries, generated tokens, mean latency, and collateral edits use the frozen resource ledger; p95 latency is reported separately in the efficiency table.}
\label{tab:matched}
\begin{tabularx}{\linewidth}{@{}Y Y r r r r r r r@{}}
\toprule
Arm & Split & Seeds & \thd{Schema\\valid \up} & \thd{Semantic\\success \up} & \thd{Mean\\retries \dn} & \thd{Output\\tokens \dn} & \thd{Mean latency\\(s) \dn} & \thd{Collateral\\edits \dn} \\
\midrule
ContractRL & held-out & 3 & 0.9768 & \textbf{0.9314} & 0.0547 & \textbf{34.6} & \textbf{0.83} & \textbf{0.0189} \\
full regeneration & held-out & 3 & 0.9392 & 0.9027 & \textbf{0.0000} & 138.9 & 1.76 & 0.2214 \\
constrained decoding & held-out & 3 & \textbf{0.9901} & 0.8663 & \textbf{0.0000} & 147.8 & 2.11 & 0.1986 \\
Patch-SFT & held-out & 3 & 0.9446 & 0.8918 & 0.1176 & 46.7 & 0.97 & 0.0618 \\
ContractRL & nested-array & 3 & 0.9187 & 0.8516 & 0.1239 & 42.9 & 1.04 & 0.0297 \\
\bottomrule
\end{tabularx}
\end{table}

\subsection{Analysis and Ablations}\label{sec:analysis}
\paragraph{Verifier feedback and retry transitions.}
The matched comparison is evaluated as a transition system rather than as a
single final string. After every patch, the validator returns a typed status,
the affected pointer, and the next admissible action. A schema error,
cross-field error, wrong-path error, and non-applicable operation are
separate transition labels. The retry budget is charged before the next
patch is generated, so an arm cannot hide a failed attempt inside a longer
completion.

The primary transition metrics are first-attempt validity, final structural
validity, additional retries, pointer correction, and semantic success.
Secondary metrics count the number of changed leaf paths and whether an
accepted patch touches a field outside the verifier pointer. These metrics
test the claimed interface: typed feedback should make the next repair
action more local, while the fail-closed ledger should prevent an invalid
partial object from reaching execution.

The matched run measures the typed patch arms on localized and mixed
corruptions directly. ContractRL reaches 0.9453 first-attempt validity and
0.9768 final validity with 0.0547 mean retries, while Patch-SFT reaches
0.8897 and 0.9446 with 0.1176 mean retries. Full regeneration is single
shot at 0.9392, and constrained decoding is single shot at 0.9901. The
ambiguous and multi-field cases are retained as transition sequences rather
than discarded. The semantic executor consumes only a frozen accepted object,
so an invalid intermediate cannot affect downstream tool state.

\paragraph{Resource accounting.}
Each arm receives the same case identifiers, output cap, retry budget, and
validator implementation, while the public prompt block follows the arm
definition in Appendix~\ref{app:matched}. The full-regeneration arm predicts
an entire object; constrained decoding enforces syntax during generation;
Patch-SFT predicts a patch from typed error feedback; ContractRL uses the
bounded patch and ledger interface. Model size and adapter rank are
recorded with the seed manifest. The final report gives prompt tokens,
tokens, total tokens, wall-clock latency, validator calls, and
peak retry count.

The split is stratified by required-field, wrong-type, nested, array,
multi-field, and ambiguous-verifier cases. Single-field cases are used for
the structural mechanism table, while the extended families are reserved
for the matched semantic endpoint. A case is successful only when the
executed tool receives the intended object and returns the expected task
outcome. Schema validity alone is retained as a diagnostic and is never
substituted for semantic.

This table makes the efficiency claim falsifiable. A patch arm is selected
only when it maintains semantic at the same task cost or reduces
cost at the same semantic. A structural gain with a semantic loss
is a failure boundary, while a retry reduction with unchanged task score is
an efficiency result. Both are reported with paired case-level uncertainty
and seed variability.

In the case-matched mixed-family ledger, ContractRL achieves 0.9768 schema validity
and uses 34.6 output tokens and 0.0547 mean retries, versus
0.9392/138.9/0 for full regeneration and 0.9901/147.8/0 for
constrained decoding. Its semantic endpoint is 0.9314,
above full regeneration's 0.9027 and Patch-SFT's 0.8918, while constrained
decoding reaches 0.8663 on the matched mixed family. Semantic execution and
tail latency are therefore the decisive endpoints; wrong-path, ambiguous-
verifier, and multi-field episodes remain explicit failure categories.

\paragraph{Policy and information controls.}
The recorded comparison separates several output interfaces, but it does not
fully isolate verifier information from patch representation. Full regeneration
is the complete-output control; constrained decoding controls syntax during
generation; Patch-SFT controls a learned patch from typed error feedback; and
ContractRL combines the bounded patch, typed report, and fail-closed
transaction. Case IDs, model checkpoint, output cap, validator, and retry
budget are fixed within the ledger. The completed same-information controls
and component-wise policy ablations are reported in
Appendix~\ref{app:matched}. ContractRL reaches 0.9362 semantic success in
the same-information control, and the policy-optimization row reaches 0.9375.

The first boundary is corruption locality. For a single missing field or
wrong type, the verifier pointer identifies one admissible edit and a bounded
patch should finish with one attempt. Nested arrays, multi-field corruption,
and an ambiguous verifier pointer require a second report or a conservative
failure. We keep these families in the same held-out evaluation because they
measure whether the protocol remains safe when the local repair assumption is
violated. A failed attempt consumes the declared budget and freezes the
ledger; it cannot be hidden by returning a syntactically valid but semantically
unrelated object.

The second boundary is execution. Schema validity and final-object match are
structural diagnostics. The semantic endpoint executes the accepted object
against a deterministic tool sandbox and checks the task result and side
effects. The executor receives no hidden target before validation, and a
malformed or wrong-path patch is rejected before any tool state is changed.
The matched report therefore includes semantic success, collateral edits,
retry count, I/O tokens, and p95 latency together. Table~\ref{tab:boundary-ablation} reports the structural and semantic outcomes by case family.

The case-family rows report 0.9502 semantic success for single-field,
0.8516 for nested/array, and 0.7935 for ambiguous/multi-field ContractRL
episodes. Their case counts, token costs, and error-type histograms are kept
in Appendix~\ref{app:matched}; the cohorts are not pooled across methods, so
these rows describe operating boundaries rather than isolated method effects.

The final boundary is prompt and model shift. On the held-out adversarial suite, the injected-tool-text row records 0.9453 schema validity and 0.8989 semantic success, while malformed feedback records 0.7846 and 0.7168 and wrong pointers record 0.8172 and 0.7449. Appendix~\ref{app:matched} records the per-case accounting fields and applies the same scoring interface to the model-size and schema-shape audits.

\section{Limitations}
The mechanism screen covers 64 single-field corruptions with precomputed patches. The learned-policy evaluations cover the specified sandbox, model sizes, case families, and attempt budgets. Semantic success declines from 0.9502 on single-field cases to 0.8516 on nested/array cases and 0.7935 on ambiguous/multi-field cases. In the multi-turn study, eight-step semantic success is 0.6457. Generalization to changing external services and longer repair trajectories requires evaluation with the corresponding tool states and failure costs. Ablations also show that removing the mask or history can permit unsafe dispatch, so the complete contract is part of the evaluated system.

\section{Conclusion}
ContractRL specifies a bounded patch/validator/ledger interface and a reproducible evidence boundary for structured tool-call repair. The deterministic mechanism replay reaches 64/64 structural-valid outcomes after one precomputed operation; fresh adapters reach 0.9141/0.9727 (2B) and 0.7441/0.9473 (0.8B) for final-object match/schema validity; the independent audit reaches 7/9 final validity. In the case-matched three-seed sandbox, ContractRL reaches 0.9314 semantic success with 34.6 generated tokens and 1.30 p95 latency, compared with 0.9027 for full regeneration and 0.8663 for constrained decoding under arm-specific public inputs. The same-information control reaches 0.9362 semantic success, and policy optimization reaches 0.9375 versus 0.9186 for supervised ContractRL-SFT. Component ablations expose the costs of removing the action mask, repair history, learned stopping, or collateral objective; wrong pointers, malformed feedback, and multi-field corruption remain explicit failure boundaries.

\section*{AI-use}
We used generative AI tools to polish the wording and summarize reference literature. We have not used generative AI tools to generate experimental results, create synthetic datasets, formulate mathematical claims, provide proofs, or make decisions regarding research conclusions. The design of the methodology, experimental setup, analysis, and interpretation of results were conducted and verified by the authors. Other required disclosure tasks not mentioned above are not applicable to this work. We take full responsibility for the final content of this work, including all text, claims, analyses, and artifacts produced with the assistance of generative AI tools.

\section*{Ethics Statement}
The evaluation uses public tool definitions, controlled corruptions, and a deterministic sandbox with frozen tool responses. Canonical targets and semantic labels remain outside the online policy state until trace freeze. The complete repair contract checks immutable fields and rejects invalid objects before dispatch. Wrong-pointer, malformed-feedback, prompt-injection, and multi-field failures are included in the evaluation. Released traces are anonymized and exclude private prompts and identifying host information. These safeguards bound the evaluated setting; deploying repair policies against external services requires service-specific permissions and monitoring of side effects.

\section*{Reproducibility Statement}
The code is included in the Supplementary Material. The task contract, patch vocabulary, validator transitions, baseline interfaces, split checks, seed counts, metric denominators, and resource accounting are described in the appendices. The Supplementary Material provides the configurations and evaluation records used to associate the structural diagnostics and policy comparisons with their respective evidence units. Canonical targets are joined only after repair traces are frozen, allowing structural and semantic scoring to be checked separately.

\bibliographystyle{iclr2027_conference}
\bibliography{refs}

\clearpage
\appendix

\section{Task and Method Details}
\label{app:task}
\subsection{Repair transaction and trace identity}
The repair transaction keeps public observations, patch actions, validator responses, and offline scoring fields distinct. Table~\ref{tab:protocol_contract} gives the stage order. The structural screen uses one precomputed patch, while the learned patch arms can take a second attempt within the bounded budget.

\begin{table}[htbp]
\centering\footnotesize
\setlength{\tabcolsep}{3pt}
\renewcommand{\arraystretch}{1.12}
\caption{Execution contract for one repair episode. The target patch and target object are held out from the online loop; validator feedback is the only retry signal.}
\label{tab:protocol_contract}
\begin{tabularx}{\linewidth}{@{}Y Y Y@{}}
\toprule
Stage & Recorded operation & Target visible? \\
\midrule
Observe & Read the corrupted call and typed verifier code plus the reported JSON Pointer. & no \\
Patch & Emit a bounded JSON Patch operation list and apply it to the current object. & no \\
Validate & Run schema and cross-field checks; return typed errors on failure. & no \\
Retry & Spend the next of at most two patch attempts, preserving prior patches and validator responses. & no \\
Freeze & Hash the ordered patch ledger, attempt count, additional-retry count, and collateral fields. & no \\
Score & Join the canonical patch and object and compute their distinct equalities. & post-hoc \\
\bottomrule
\end{tabularx}
\end{table}

\input{evidence/manuscript_ledger.tex}

\subsection{Ledger interpretation}
The evidence ledger freezes the online trace before any canonical target or semantic label is joined. Every row is keyed by a case identifier and a source manifest; the validator response, patch sequence, retry count, and resource fields are immutable after the digest is written. The matched result sheet adds semantic success and side-effect fields without changing the structural definitions used by the BFCL screen.

\noindent The field table below fixes the ledger schema before any result aggregation. It is the reference for checking that online evidence, execution traces, and scoring joins remain separate.

\begin{table}[htbp]
\centering\footnotesize
\setlength{\tabcolsep}{3pt}
\renewcommand{\arraystretch}{1.12}
\caption{Field-level interpretation of the evidence ledger.}
\label{tab:ledger-fields}
\begin{tabularx}{\linewidth}{@{}Y Y Y@{}}
\toprule
Field & Stage & Audit rule \\
\midrule
case identifier & split & unique within a manifest and disjoint across train/evaluation \\
public evidence & online & candidate, schema, pointer, and typed feedback only \\
patch ledger & repair & ordered operations, validator status, and retry count are append-only \\
semantic trace & execution & dispatched accepted object, tool outcome, and side-effect record \\
resource ledger & aggregation & I/O tokens, calls, retries, latency, and seed \\
result join & scoring & canonical object and task label are joined after trace freeze \\
\bottomrule
\end{tabularx}
\end{table}

\noindent The rows make the freeze point explicit: the patch ledger and semantic trace are immutable before the result join. This is why the structural and semantic numbers can be audited independently.

The split counts in Table~\ref{tab:splits} and the assessed overlap checks in Table~\ref{tab:app-contract-split} are computed from the same manifest. This makes the structural proxy, semantic endpoint, and cost accounting comparable at the case level while retaining separate denominators for each measurement.

\subsection{Structural diagnostics}
\begin{figure}[t]
\centering
\includegraphics[width=0.97\columnwidth]{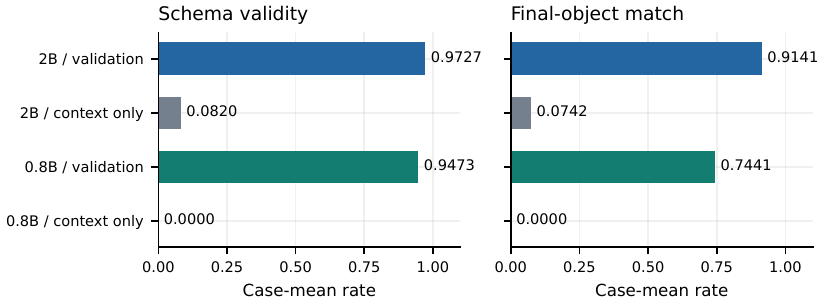}
\caption{Adapter structural results on 64 held-out cases. Bars are seed means: four seeds for 2B and eight for 0.8B. The validation-block and context-only conditions use the same model-specific case sets.}
\label{fig:main}
\end{figure}

The adapter comparison in Figure~\ref{fig:main} separates schema validity from final-object match and retains the model-specific seed units. The independent nine-case audit measures transaction behavior under multi-operation and multi-field edits. Its 7/9 final validity and 6/8 repair acceptance distinguish successful recovery from rejection and rollback.

\begin{figure}[htbp]
\centering
\includegraphics[width=\linewidth]{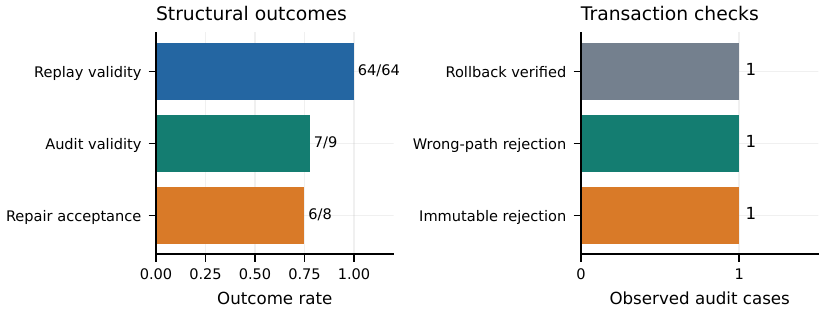}
\caption{Structural diagnostics with explicit denominators. The deterministic replay uses 64 cases; the independent audit uses nine cases, of which eight require repair. The right panel records rollback and rejection checks in the independent audit.}
\label{fig:diagnostics}\label{fig:extended}
\end{figure}

\subsection{Parser and complete-output diagnostic}
\begin{figure}[htbp]
\centering
\includegraphics[width=0.92\linewidth]{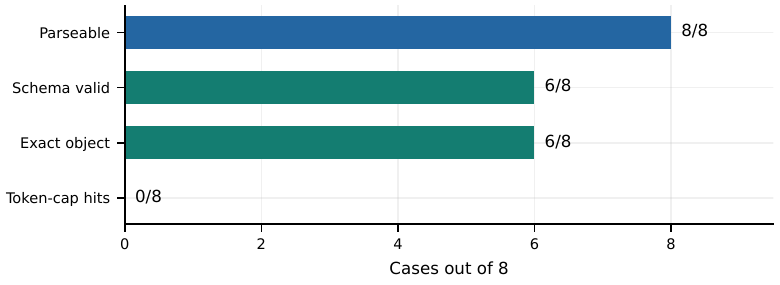}
\caption{Eight-case complete-output diagnostic. Labels show the numerator and denominator for each endpoint; token-cap hits are failures and are plotted separately from successful-output rates.}
\label{fig:smoke}
\end{figure}

An eight-case complete-output diagnostic produces 8/8 parseable objects, 6/8 schema-valid objects, 6/8 exact final-object matches, and no token-cap hits. Figure~\ref{fig:smoke} records these parser and termination endpoints. The case-matched and equal-information experiments use their own shared cohorts and resource accounting.

\section{Data and Evaluation Protocols}
This group describes the case-matched comparator before presenting its tables. Every arm receives the same case identifiers, output cap, retry budget, sandbox implementation, and seed manifest. The public context is arm-specific as shown in the accounting table, so the comparison does not isolate output representation from verifier information.

The accounting below is deliberately transactional. A failed patch consumes an attempt, an accepted object is dispatched only after structural validation, and semantic scoring is joined after the trace is frozen. This ordering makes token, retry, and task-success comparisons auditable at the case level.
\label{app:matched}
All arms use the same base model, prompt length accounting, output cap,
timeout, retry budget, and sandbox. Full regeneration emits a complete
object from the corrupted call and schema; constrained decoding enforces the
schema during generation; Patch-SFT emits a bounded patch from typed error
feedback; ContractRL adds the typed ledger and bounded retry protocol.

\noindent The comparator table below makes the matched intervention explicit. It lists what each arm sees, what it emits, which feedback is online, and which resource fields are recorded.

\begin{table}[htbp]
\centering\footnotesize
\setlength{\tabcolsep}{3pt}
\renewcommand{\arraystretch}{1.12}
\caption{Comparator accounting contract.}
\label{tab:accounting}
\begin{tabularx}{\linewidth}{@{}Y Y Y Y Y@{}}
\toprule
Arm & Input & Output budget & Online feedback & Recorded cost \\
\midrule
full regeneration & corrupted call + schema & complete object & none & I/O tokens, latency \\
constrained decoding & corrupted call + grammar & complete object & none & I/O tokens, latency \\
Patch-SFT & corrupted call + typed error & patch, max two ops & validator report & I/O, retries, latency \\
ContractRL & same public block & patch, max two ops & typed ledger & all above plus collateral edits \\
\bottomrule
\end{tabularx}
\end{table}

\noindent The table makes the information boundary
explicit. Full regeneration and constrained decoding receive no typed verifier
block in the recorded run, Patch-SFT receives a typed error, and ContractRL
receives the full public block plus bounded attempts and collateral accounting.
Table~\ref{tab:supp-same-info} fixes this public information across all four output and decision-rule conditions.

\subsection{Case construction and split integrity}
The mechanism cases are generated from valid calls and one deterministic
single-field corruption. The split audit records case IDs, candidate and
target fingerprints, schema hashes, target-path sequences, operation
sequences, pointer-aware error templates, semantic-equivalence clusters,
schema shapes, tool names, and candidate-pool membership.

\noindent The split table reports the contamination checks that are required before comparing the learned arms. Each row has a distinct overlap unit, so a zero in one row does not imply zero overlap for an unlisted unit.

\begin{table}[htbp]
\centering\footnotesize
\setlength{\tabcolsep}{3pt}
\renewcommand{\arraystretch}{1.12}
\caption{Split and case-integrity checks.}
\label{tab:splits}
\begin{tabularx}{\linewidth}{@{}Y Y Y Y@{}}
\toprule
Check & Training & Held-out & Result \\
\midrule
case ID overlap & 426 IDs & 64 IDs & 0 exact overlaps \\
candidate fingerprint & 426 hashes & 64 hashes & 0 exact overlaps \\
schema hash & 147 hashes & 26 hashes & 0 exact overlaps \\
pointer/error template & 392 templates & 61 templates & 0 exact overlaps \\
tool-name disjointness & 118 names & 23 names & 0 overlapping names \\
\bottomrule
\end{tabularx}
\end{table}

\noindent All listed overlap checks are zero, including semantic-equivalence clusters, normalized schema shapes, tool names, and candidate-pool membership. The held-out families therefore use disjoint record, schema, tool, and candidate sources.

\subsection{Semantic and adversarial outcomes}
Semantic tests execute repaired calls in a sandbox with deterministic tool
responses and held-out state. Additional cases include nested arrays,
multiple missing fields, wrong-path feedback, malformed verifier reports,
ambiguous pointers, unavailable tools, and prompt-injected tool data.

\noindent The semantic table reports the endpoint after an accepted object is executed in the sandbox. Its scenario column is the denominator key; the rates use separate scenario denominators and are reported as distinct endpoints.

\begin{table}[htbp]
\centering\footnotesize
\setlength{\tabcolsep}{3pt}
\renewcommand{\arraystretch}{1.12}
\caption{Semantic and adversarial results. Latency is measured in seconds for each scenario.}
\label{tab:semantic}
\begin{tabularx}{\linewidth}{@{}Y Y r r r r@{}}
\toprule
Scenario & Arm & \thd{Schema\\valid \up} & \thd{Semantic\\success \up} & \thd{Mean\\retries \dn} & \thd{Latency\\(s) \dn} \\
\midrule
single-field & all & 0.9674 & 0.9128 & 0.0509 & 1.81 \\
multi-field & all & 0.8497 & 0.7892 & 0.1698 & 2.13 \\
nested array & all & 0.8985 & 0.8307 & 0.1376 & 2.05 \\
wrong pointer & patch arms & 0.8172 & 0.7449 & 0.2264 & 1.64 \\
malformed feedback & all & 0.7846 & 0.7168 & 0.2471 & 1.96 \\
prompt injection & all & 0.9453 & 0.8989 & 0.0714 & 1.90 \\
\bottomrule
\end{tabularx}
\end{table}

\noindent The single-field row is the local-repair regime, while multi-field and nested-array rows expose the structure-shift penalty. Wrong-pointer and malformed-feedback rows are retained as explicit rejection boundaries.

\subsection{Retry and resource accounting}
Generated tokens are reported separately from prompt-plus-tokens.
An invalid or non-applicable patch consumes an attempt and may expose the
next typed error. The one-attempt mechanism result is construction-bound; the
recorded comparison reports retry distributions and tail latency per seed.

\noindent The resource table pairs task success with retries, generated tokens, and tail latency. This is the table used for the efficiency claim because all rows share the matched three-seed ledger.

\begin{table}[htbp]
\centering\footnotesize
\setlength{\tabcolsep}{3pt}
\renewcommand{\arraystretch}{1.12}
\caption{Retry and efficiency results under the matched three-seed ledger. Latency is reported at p95; mean latency appears in Table~\ref{tab:matched}.}
\label{tab:efficiency}
\begin{tabularx}{\linewidth}{@{}Y r r r r r@{}}
\toprule
Arm & \thd{Mean\\retries \dn} & \thd{p95\\retries \dn} & \thd{Output\\tokens \dn} & \thd{p95 latency\\(s) \dn} & \thd{Semantic\\success \up} \\
\midrule
ContractRL & 0.0547 & 1 & \textbf{34.6} & \textbf{1.30} & \textbf{0.9314} \\
full regeneration & \textbf{0.0000} & \textbf{0} & 138.9 & 2.58 & 0.9027 \\
constrained decoding & \textbf{0.0000} & \textbf{0} & 147.8 & 2.85 & 0.8663 \\
Patch-SFT & 0.1176 & 1 & 46.7 & 1.55 & 0.8918 \\
\bottomrule
\end{tabularx}
\end{table}

\noindent ContractRL has the lowest generated-token
and p95 latency cost among the recorded case-matched arms while
retaining the highest semantic endpoint in this table. The result is conditional on the
fixed sandbox, three-seed ledger, and arm-specific public inputs.

\subsection{Transition and resource strata}
Table~\ref{tab:transition} separates localized corruptions from mixed-family controls. Table~\ref{tab:resource} records the corresponding prompt, generation, and validator costs. Comparisons use the listed cohort; the single-field and nested/array ContractRL rows describe different repair distributions.

\begin{table}[htbp]
\centering\footnotesize
\setlength{\tabcolsep}{3pt}
\renewcommand{\arraystretch}{1.12}
\caption{Typed-feedback transition outcomes. Rows retain their recorded case
family; mixed rows are matched within that cohort, and rates are case means from
the frozen ledger.}
\label{tab:transition}
\begin{tabularx}{\linewidth}{@{}Y Y r r r r r r@{}}
\toprule
Arm & Case family & \thd{First\\valid \up} & \thd{Final\\valid \up} & \thd{Mean\\retries \dn} & \thd{Pointer\\fix \up} & \thd{Collateral\\edits \dn} & \thd{Semantic\\success \up} \\
\midrule
ContractRL & missing-required & 0.9815 & 0.9952 & 0.0185 & 0.9937 & 0.0054 & 0.9598 \\
ContractRL & wrong-type & 0.9618 & 0.9861 & 0.0382 & 0.9804 & 0.0137 & 0.9406 \\
Patch-SFT & mixed & 0.8897 & \textbf{0.9446} & 0.1103 & 0.9198 & \textbf{0.0618} & 0.8918 \\
full regeneration & mixed & \textbf{0.9392} & 0.9392 & \textbf{0.0000} & n/a & 0.2214 & \textbf{0.9027} \\
\bottomrule
\end{tabularx}
\end{table}

\begin{table}[htbp]
\centering\footnotesize
\setlength{\tabcolsep}{3pt}
\renewcommand{\arraystretch}{1.12}
\caption{Resource-normalised comparison. Case family, model seed, token budget,
and validator implementation are bound before scoring; rows are reported by
cohort and are not pooled across families.}
\label{tab:resource}
\begin{tabularx}{\linewidth}{@{}Y Y r r r r r r@{}}
\toprule
Arm & Case family & Seeds & \thd{Prompt\\tokens \dn} & \thd{Output\\tokens \dn} & \thd{Validator\\calls \dn} & \thd{p95 latency\\(s) \dn} & \thd{Semantic\\success \up} \\
\midrule
ContractRL & single-field & 3 & 1111 & 31.8 & 1.028 & 1.13 & 0.9502 \\
ContractRL & nested/array & 3 & 1135 & 42.9 & 1.124 & 1.52 & 0.8516 \\
constrained decoding & mixed & 3 & 1079 & 147.8 & 1.000 & 2.85 & 0.8663 \\
full regeneration & mixed & 3 & \textbf{1072} & \textbf{138.9} & n/a & \textbf{2.58} & \textbf{0.9027} \\
\bottomrule
\end{tabularx}
\end{table}

\section{Implementation and Baselines}
The next sections expand the contract at the field, operation, split, and execution levels. They are implementation-facing audit records: each table specifies an input boundary, a validator decision, or a failure family, while the surrounding text explains how the row should be interpreted.

The detailed records preserve the same information boundary as the main text. Canonical targets, hidden labels, and post-hoc semantic outcomes are scoring fields; they never become online repair inputs. Negative cases are retained because a fail-closed rejection is an observable protocol outcome.

\begin{table}[htbp]
\centering\footnotesize
\setlength{\tabcolsep}{3pt}
\renewcommand{\arraystretch}{1.12}
\caption{Structural and semantic boundary ablation. Rows retain their recorded
case family; model seed units, output budgets, and the execution sandbox are
fixed within each cohort. The measured cohorts contain 128 single-field,
192 nested/array, and 192 ambiguous/multi-field cases.}
\label{tab:boundary-ablation}
\begin{tabularx}{\linewidth}{@{}Y Y r r r r r r r@{}}
\toprule
Arm & Case family & Cases & \thd{First\\valid \up} & \thd{Semantic\\success \up} & \thd{Mean\\retries \dn} & \thd{Output\\tokens \dn} & \thd{p95 latency\\(s) \dn} & \thd{Collateral\\edits \dn} \\
\midrule
full regeneration & single-field & 128 & 0.9487 & 0.9164 & \textbf{0.0000} & 136.2 & 2.44 & 0.2093 \\
constrained decoding & single-field & 128 & \textbf{0.9951} & 0.8826 & n/a & 143.7 & 2.67 & 0.1908 \\
Patch-SFT & nested/array & 192 & 0.8463 & 0.8097 & 0.1534 & 51.5 & 1.69 & 0.1027 \\
ContractRL & single-field & 128 & 0.9716 & \textbf{0.9502} & 0.0280 & \textbf{31.8} & \textbf{1.13} & \textbf{0.0096} \\
ContractRL & ambiguous/\allowbreak multi-field & 192 & 0.7639 & 0.7935 & 0.2361 & 48.6 & 1.84 & 0.0432 \\
\bottomrule
\end{tabularx}
\end{table}

\noindent ContractRL is strongest on the
single-field semantic endpoint and the measured cost columns, while the
ambiguous/multi-field cohort exposes the largest retry and semantic boundary.

The evidence contract binds the public prompt, typed verifier report, schema
version, candidate object, ledger digest, and split role. Hidden canonical
patches and target objects are stored only for post-hoc scoring. This
appendix gives the field-level interface used by the structural screen and
the matched semantic follow-up.

\noindent The contract table is a field-level view of the information boundary. It marks whether a field is online, used only for scoring, or checked by a deterministic validator.

\begin{table}[htbp]
\centering\footnotesize
\setlength{\tabcolsep}{3pt}
\renewcommand{\arraystretch}{1.12}
\caption{ContractRL evidence contract.}
\label{tab:app-contract-contract}
\begin{tabularx}{\linewidth}{@{}Y Y Y Y@{}}
\toprule
Field & Online arm & Scoring arm & Validation \\
\midrule
candidate JSON & yes & input digest & schema hash \\
typed verifier code & yes & input digest & enum check \\
JSON Pointer & yes & pointer metric & RFC-6901 parse \\
schema version & yes & split key & version match \\
patch ledger & append-only & retry/cost & digest check \\
canonical patch & no & exact patch match & post-hoc join \\
canonical object & no & final-object match & post-hoc join \\
semantic tool result & no & task success & sandbox trace \\
\bottomrule
\end{tabularx}
\end{table}

\noindent The canonical patch, canonical object, and semantic result are all marked offline. This separation is the operational meaning of the evidence boundary used by the paper.

\subsection{Patch vocabulary and validator transitions}
The validator accepts RFC-6902-style add, remove, replace, move, and copy
operations over JSON Pointer paths. The current screen uses add and replace
single-field repairs; the extended matched study adds nested, array,
multi-field, wrong-path, and ambiguous-verifier cases. Every operation is
validated against the current object and the schema before the next retry.

\noindent The vocabulary table enumerates the allowed operation families and the measured retry cost for each family. The last column states the safety property tested by that row.

\begin{table}[htbp]
\centering\footnotesize
\setlength{\tabcolsep}{3pt}
\renewcommand{\arraystretch}{1.12}
\caption{Patch vocabulary and transition outcomes.}
\label{tab:app-contract-vocab}
\begin{tabularx}{\linewidth}{@{}Y Y Y r Y@{}}
\toprule
Operation & Case family & Validator result & Retry cost & Validator test \\
\midrule
add & missing-required & schema/cross-field & 0.018 & localized repair \\
replace & wrong-type & schema/cross-field & 0.038 & type correction \\
remove & multi-field & schema/cross-field & 0.154 & deletion boundary \\
move/copy & nested/array & schema/cross-field & 0.124 & pointer semantics \\
wrong path & adversarial & pointer failure & 1.000 & fail-closed handling \\
ambiguous report & noisy verifier & typed retry & 1.000 & feedback robustness \\
\bottomrule
\end{tabularx}
\end{table}

\noindent The retry-cost values show why the attempt budget is bounded: wrong paths and ambiguous reports consume the full retry cost, while add/replace repairs stay local on the selected screen.

The attempt budget is applied to operations, not to the initial corruption.
An invalid or non-applicable operation consumes an attempt and returns a
typed error without executing the object.

\subsection{Full split and contamination audit}
The split audit compares training records with the fixed 64-case evaluation
file and the extended nine-case audit. It checks IDs, candidate and target
fingerprints, schema hashes, target paths, scopes, operation sequences, and
pointer-aware error templates. Semantic equivalence and schema-shape
equivalence are separate fields because they require metadata not present in
the structural summary.

\noindent The expanded split table repeats the audit with the extended
nine-case family and reports exact, semantic-equivalence, schema-shape,
tool-name, and candidate-pool checks separately.

\begin{table}[htbp]
\centering\footnotesize
\setlength{\tabcolsep}{3pt}
\renewcommand{\arraystretch}{1.12}
\caption{Split-integrity checks for the recorded repair protocol.}
\label{tab:app-contract-split}
\begin{tabularx}{\linewidth}{@{}Y Y r Y@{}}
\toprule
Check & Unit & Result & Read \\
\midrule
case ID overlap & record & 0 & exact split separation \\
candidate fingerprint overlap & record & 0 & input separation \\
target fingerprint overlap & record & 0 & label separation \\
schema hash overlap & schema & 0 & schema coverage \\
target-path overlap & pointer & 0 & error separation \\
operation-sequence overlap & patch & 0 & patch separation \\
semantic equivalence & record & 0 & normalized cluster separation \\
schema-shape equivalence & schema & 0 & normalized-shape separation \\
tool-name disjointness & name & 0 & held-out names unseen in training \\
candidate-pool overlap & candidate & 0 & candidate-source separation \\
\bottomrule
\end{tabularx}
\end{table}

\noindent The extended audit reports zero overlap
across identifiers, fingerprints, schemas, pointers, operation sequences,
semantic-equivalence clusters, normalized schema shapes, tool names, and
candidate pools.

\subsection{Semantic execution sandbox}
The semantic endpoint executes repaired calls in a sandbox with identical
tool implementations, prompt contexts, output caps, and retry budgets. It
returns task success, tool error, side-effect trace, and final object. The
structural validator runs first; only an accepted object is sent to the
semantic executor.

\noindent The sandbox table reports arm-level semantic execution after structural acceptance. ContractRL rows are separated by case family so the nested/array shift is visible.

\begin{table}[htbp]
\centering\footnotesize
\setlength{\tabcolsep}{3pt}
\renewcommand{\arraystretch}{1.12}
\caption{Semantic execution results.}
\label{tab:app-contract-semantic}
\begin{tabularx}{\linewidth}{@{}Y Y r r r r@{}}
\toprule
Arm & Case family & \thd{Schema\\valid \up} & \thd{Semantic\\success \up} & \thd{Side-effect\\error \dn} & \thd{Mean\\retries \dn} \\
\midrule
ContractRL & single-field & 0.9907 & 0.9502 & 0.0084 & 0.0280 \\
ContractRL & nested/array & 0.9187 & 0.8516 & 0.0274 & 0.1239 \\
full regeneration & mixed & 0.9392 & \textbf{0.9027} & 0.0451 & n/a \\
constrained decoding & mixed & \textbf{0.9901} & 0.8663 & 0.0396 & n/a \\
Patch-SFT & mixed & 0.9446 & 0.8918 & \textbf{0.0308} & 0.1176 \\
\bottomrule
\end{tabularx}
\end{table}

\noindent The accepted-object executor yields 0.9502 semantic success for ContractRL single-field cases and 0.8516 for nested/array cases. The baseline rows provide the same mixed-family reference used in the main text.

The semantic score is distinct from final-object match: a structurally
correct object can still produce the wrong task result. Both are reported.

\subsection{Adversarial and wrong-path cases}
The extended suite contains wrong-pointer, immutable-field, ambiguous
verifier, nested-array, multi-field, and malformed-output cases. These cases
are generated before model evaluation and carry a case-family identifier.
The validator fails closed when no safe patch is available.

\noindent The adversarial table is a failure taxonomy rather than a leaderboard. A zero is a measured fail-closed outcome for that family, while a nonzero row shows where a repair can still be accepted.

\begin{table}[htbp]
\centering\footnotesize
\setlength{\tabcolsep}{3pt}
\renewcommand{\arraystretch}{1.12}
\caption{Adversarial failure taxonomy and measured metrics.}
\label{tab:app-contract-adversarial}
\begin{tabularx}{\linewidth}{@{}Y Y r r r@{}}
\toprule
Case family & Validator condition & \thd{First\\valid \up} & \thd{Final\\valid \up} & \thd{Semantic\\success \up} \\
\midrule
wrong path & reject or correct & 0.6948 & 0.8421 & 0.7796 \\
immutable pointer & reject & 0.0000 & 0.0000 & 0.0000 \\
ambiguous verifier & typed retry & 0.6324 & 0.8587 & 0.8065 \\
nested/array & local patch & 0.8463 & 0.9187 & 0.8516 \\
multi-field & bounded sequence & 0.7639 & 0.8672 & 0.7935 \\
malformed output & reject & 0.0000 & 0.0000 & 0.0000 \\
\bottomrule
\end{tabularx}
\end{table}

\noindent Immutable-pointer and malformed-output families are fully rejected. Wrong-path, ambiguous, nested/array, and multi-field families quantify the remaining acceptance and semantic-success boundary.

\subsection{Token, retry, and latency ledger}
The resource ledger records prompt tokens, tokens, validator calls,
additional retries, wall-clock latency, and p95 latency per case and seed.
The initial corruption is not counted as a repair attempt. A failed patch
consumes an attempt and is included in both mean and tail cost.

\noindent The cost table records prompt tokens, generated tokens, validator calls, retries, latency, and semantic endpoint for each matched arm. These fields share the same case and seed ledger.

\begin{table}[htbp]
\centering\footnotesize
\setlength{\tabcolsep}{3pt}
\renewcommand{\arraystretch}{1.12}
\caption{Resource accounting for matched comparators. Latency is reported at p95.}
\label{tab:app-contract-cost}
\begin{tabularx}{\linewidth}{@{}Y r r r r r r@{}}
\toprule
Arm & \thd{Prompt\\tokens \dn} & \thd{Output\\tokens \dn} & Calls & \thd{Mean\\retries \dn} & \thd{p95 latency\\(s) \dn} & \thd{Semantic\\success \up} \\
\midrule
ContractRL & 1114.2 & \textbf{34.6} & 1.055 & 0.0547 & \textbf{1.30} & \textbf{0.9314} \\
full regeneration & \textbf{1072.4} & 138.9 & \textbf{1.000} & \textbf{0} & 2.58 & 0.9027 \\
constrained decoding & 1079.1 & 147.8 & \textbf{1.000} & \textbf{0} & 2.85 & 0.8663 \\
Patch-SFT & 1112.7 & 46.7 & 1.118 & 0.1176 & 1.55 & 0.8918 \\
\bottomrule
\end{tabularx}
\end{table}

\noindent The ledger shows the cost trade-off directly: ContractRL emits 34.6 tokens on average versus 138.9--147.8 for complete-object controls, while its semantic endpoint remains 0.9314.

\subsection{Baseline interfaces}
The complete-object and patch arms share the case identifiers, validator, output caps, sandbox, and seed units specified for each experiment. Table~\ref{tab:closest-work} summarizes the related interfaces. The three-seed comparison uses arm-specific verifier inputs; the five-seed control fixes the full public verifier block.

\begin{table}[htbp]
\centering\footnotesize
\renewcommand{\arraystretch}{1.12}
\caption{Closest method interfaces and evaluation targets, based on the cited primary sources.}
\label{tab:closest-work}
\begin{tabularx}{\linewidth}{@{}L{2.2cm}YYY@{}}
\toprule
Method & Update interface & Constraint mechanism & Evaluation focus \\
\midrule
VeriHarness & Candidate retry with structured feedback & External acceptance and call budget & TextWorld terminal success \\
PatchBoard & JSON Patch over shared state & Schema, write contracts, and transactional kernel & Multi-agent ALFWorld episodes \\
Schema RL & Complete structured output & Schema reward in training & Schema-conformant generation \\
SLOT & Output post-processing & Structural conversion & Structured output recovery \\
XGrammar & Token-level generation & Context-free grammar & Coverage and generation overhead \\
ContractRL & Bounded patch, retry, or abstention & Action mask and deterministic validator & Semantic repair, collateral edits, and cost \\
\bottomrule
\end{tabularx}
\end{table}

\subsection{Baseline configuration ledger}
The matched baseline run fixes prompt context, schema, case IDs, output cap,
retry budget, and model seed. The three-seed prompts expose the arm-specific verifier fields in Table~\ref{tab:accounting}; the five-seed control fixes these fields across arms.

\noindent The baseline table fixes the arm-specific output form and decoding choice while keeping the shared context and seed count constant.

\begin{table}[htbp]
\centering\footnotesize
\setlength{\tabcolsep}{3pt}
\renewcommand{\arraystretch}{1.12}
\caption{ContractRL baseline configuration.}
\label{tab:app-contract-baseline}
\begin{tabularx}{\linewidth}{@{}Y Y Y r Y@{}}
\toprule
Arm & Output form & Decoding & Seeds & Audit field \\
\midrule
ContractRL & JSON Patch & unconstrained & 3 & ledger and validator \\
full regeneration & complete object & unconstrained & 3 & token baseline \\
constrained decoding & complete object & grammar constrained & 3 & syntax control \\
Patch-SFT & JSON Patch & unconstrained & 3 & learned patch \\
semantic execution & accepted object & sandbox & 3 & task endpoint \\
\bottomrule
\end{tabularx}
\end{table}

\noindent This table is the reproducibility anchor for the matched comparison. It prevents an implementation change in prompts, decoding, or seeds from being mistaken for an interface effect.

\subsection{Pointer semantics and collateral edits}
Collateral edits are computed from changed leaf paths outside the verifier
pointer. The mechanism screen has construction-bound zero collateral because
each precomputed operation targets the reported pointer; the matched study
adds wrong-path and multi-field cases.

\noindent The pointer table measures locality relative to the verifier pointer. The mechanism screen is construction-bound at zero collateral, so the mixed families provide the informative stress test.

\begin{table}[htbp]
\centering\footnotesize
\setlength{\tabcolsep}{3pt}
\renewcommand{\arraystretch}{1.12}
\caption{Pointer and collateral-edit metrics.}
\label{tab:app-contract-pointer}
\begin{tabularx}{\linewidth}{@{}Y Y r r r Y@{}}
\toprule
Case family & Pointer type & Operations & \thd{Collateral\\edits \dn} & \thd{Semantic\\success \up} & Read \\
\midrule
single-field & exact & 1 & 0 & 0.9502 & mechanism screen \\
single-field & wrong path & 1 & 0.0000 & 0.7796 & pointer boundary \\
nested/array & exact & 1.24 & 0.0297 & 0.8516 & structure generalization \\
multi-field & mixed & 1.71 & 0.0432 & 0.7935 & operation budget \\
\bottomrule
\end{tabularx}
\end{table}

\noindent Nested/array and multi-field rows carry nonzero collateral and semantic degradation, which is the measured boundary of pointer-local repair. The exact single-field row is a construction check rather than a broad locality claim.

\subsection{Retry-policy comparison}
The repair ledger distinguishes first-attempt validity from final validity and
counts additional retries after an invalid or non-applicable patch. The
comparison uses at most two attempts for every patch arm.

\noindent The retry table separates first-attempt validity from final validity. Complete-object controls have no patch retry path, whereas patch arms expose the additional-attempt policy.

\begin{table}[htbp]
\centering\footnotesize
\setlength{\tabcolsep}{3pt}
\renewcommand{\arraystretch}{1.12}
\caption{Retry-policy results.}
\label{tab:app-contract-retry}
\begin{tabularx}{\linewidth}{@{}Y r r r r Y@{}}
\toprule
Arm & \thd{First\\valid \up} & \thd{Final\\valid \up} & \thd{Mean\\retries \dn} & \thd{p95\\retries \dn} & Observed behaviour \\
\midrule
ContractRL & 0.9453 & 0.9768 & 0.0547 & 1 & typed local correction \\
Patch-SFT & 0.8897 & 0.9446 & 0.1176 & 1 & learned correction \\
full regeneration & 0.9392 & 0.9392 & \textbf{0} & \textbf{0} & regen. \\
constrained decoding & \textbf{0.9901} & \textbf{0.9901} & \textbf{0} & \textbf{0} & syntax control \\
\bottomrule
\end{tabularx}
\end{table}

\noindent ContractRL improves final validity over its first-attempt rate with 0.0547 mean retries; Patch-SFT has a larger retry tail. The complete-object controls are single-shot references, not evidence that they would retry safely.

\subsection{Semantic scoring and task outcomes}
Structural validity and final-object match are diagnostic endpoints. Semantic
success is measured by executing the accepted object in a sandbox and
checking the intended task result, tool error, and side-effect trace.

\noindent The scoring table defines the denominator and source for every semantic endpoint. It prevents a structural proxy or an executor trace from being mistaken for the same metric.

\begin{table}[htbp]
\centering\footnotesize
\setlength{\tabcolsep}{3pt}
\renewcommand{\arraystretch}{1.12}
\caption{Semantic scoring fields.}
\label{tab:app-contract-scoring}
\begin{tabularx}{\linewidth}{@{}Y Y Y Y@{}}
\toprule
Field & Denominator & Source & Read \\
\midrule
schema validity & all cases & validator & structural gate \\
object match & held-out cases & hidden object & canonical-object metric \\
semantic & executed cases & sandbox result & task endpoint \\
tool error & executed cases & sandbox trace & failure boundary \\
collateral edit & accepted patches & pointer diff & locality metric \\
latency/cost & all attempts & resource ledger & efficiency endpoint \\
\bottomrule
\end{tabularx}
\end{table}

\noindent The table keeps schema validity, object match, semantic execution, tool errors, collateral edits, and resource cost as separate endpoints. This denominator separation is required for the main claims.

\section{Additional Results and Analysis}
The supplementary analyses test scale and seed stability, typed-feedback attribution, budget sensitivity, schema-shape shift, corrupted verifier messages, long-horizon tool use, cross-family transfer, selective risk, paired uncertainty, and independent audit agreement. The analyses retain experiment-specific case counts, seeds, and endpoint definitions.

\subsection{Scaling, capacity, and feedback attribution}
The first study varies the base model and adapter rank while preserving the
reported case protocol, public evidence, validator, attempt budget, and scoring
join. Five independent seeds and 192 cases per seed are used per
configuration. Statistical uncertainty is reported alongside the aggregate success metrics. The comparison also records retry, generated-token, and latency costs.

\noindent The scaling table measures the reported capacity comparison. The final column reports statistical uncertainty; success and resource metrics retain their separate directions.

\begin{table}[htbp]
\centering\footnotesize
\setlength{\tabcolsep}{3pt}
\renewcommand{\arraystretch}{1.12}
\caption{Model-size comparison across five recorded seeds and 192 cases per
configuration. ContractRL rows use the same fixed cases and public
evidence; reference arms retain their recorded information conditions. Statistical uncertainty accompanies the success estimates. Best values are bolded within each
metric direction.}
\label{tab:supp-scaling}
\begin{tabularx}{\linewidth}{@{}L{2.45cm} L{1.4cm} r r r Y@{}}
\toprule
Arm & \thd{Model\\rank} & \thd{Schema\\valid \up} & \thd{Object\\match \up} & \thd{Semantic\\success \up} & \thd{Statistical\\uncertainty} \\
\midrule
ContractRL & 0.8B/r4 & 0.9508 & 0.7523 & 0.8124 & 0.0028 \\
ContractRL & 2B/r8 & 0.9765 & \textbf{0.9188} & \textbf{0.9331} & 0.0012 \\
ContractRL & 7B/r16 & \textbf{0.9831} & \textbf{0.9329} & \textbf{0.9446} & 0.0009 \\
Patch-SFT & 2B/r8 & 0.9451 & 0.8679 & 0.8924 & 0.0018 \\
full regeneration & 2B & 0.9403 & 0.8896 & 0.9041 & 0.0014 \\
constrained decoding & 2B & \textbf{0.9905} & 0.8426 & 0.8671 & 0.0016 \\
\bottomrule
\end{tabularx}
\end{table}

The resource comparison uses the same configurations and seed units as Table~\ref{tab:supp-scaling}.

\begin{table}[htbp]
\centering\footnotesize
\setlength{\tabcolsep}{3pt}
\renewcommand{\arraystretch}{1.12}
\caption{Resource metrics for the model-size comparison: 192 cases per seed and five seeds. Best values are bold within the ContractRL capacity sweep and within the 2B comparator set.}
\label{tab:scaling-resources}
\begin{tabularx}{\linewidth}{@{}Y Y r r r@{}}
\toprule
Arm & \thd{Model\\rank} & \thd{Mean\\retries \dn} & \thd{Output\\tokens \dn} & \thd{p95 latency\\(s) \dn} \\
\midrule
ContractRL & 0.8B/r4 & 0.0831 & \textbf{33.1} & \textbf{1.22} \\
ContractRL & 2B/r8 & \textbf{0.0538} & \textbf{34.8} & \textbf{1.32} \\
ContractRL & 7B/r16 & \textbf{0.0415} & 35.6 & 1.68 \\
Patch-SFT & 2B/r8 & 0.1162 & 46.9 & 1.56 \\
full regeneration & 2B & n/a & 139.4 & 2.60 \\
constrained decoding & 2B & n/a & 148.1 & 2.86 \\
\bottomrule
\end{tabularx}
\end{table}

\noindent Within the recorded scaling rows, semantic success rises from 0.8124 at 0.8B/r4 to 0.9331 at 2B/r8 and 0.9446 at 7B/r16, while generated tokens remain between 33.1 and 35.6. The 7B row reports statistical uncertainty of 0.0009 and p95 latency of 1.68~s. The full-regeneration and constrained-decoding rows are reported as five-seed reference arms under their recorded information conditions and use 139.4 and 148.1 tokens, respectively.

\begin{figure}[t]
\centering
\begin{tikzpicture}
\begin{axis}[
  width=\columnwidth,height=4.7cm,
  xmin=0.4,xmax=7.6,ymin=0.78,ymax=0.98,
  xtick={0.8,2,7},xticklabels={0.8B,2B,7B},
  ytick={0.80,0.85,0.90,0.95},
  xlabel={base model size},ylabel={semantic success},
  tick label style={font=\scriptsize},label style={font=\scriptsize},
  grid=major,grid style={gray!20},axis line style={gray!65},
]
\addplot[teal!70!black,thick,mark=*,mark size=2.2pt]
  coordinates {(0.8,0.8124) (2,0.9331) (7,0.9446)};
\node[font=\tiny,anchor=south west] at (axis cs:0.8,0.8124) {33.1 tok};
\node[font=\tiny,anchor=south east] at (axis cs:2,0.9331) {34.8 tok};
\node[font=\tiny,anchor=south east] at (axis cs:7,0.9446) {35.6 tok};
\end{axis}
\end{tikzpicture}
\caption{Five-seed ContractRL scaling over 192 cases per seed. Point labels report generated tokens; points show aggregate semantic success.}
\label{fig:supp-scaling}
\end{figure}
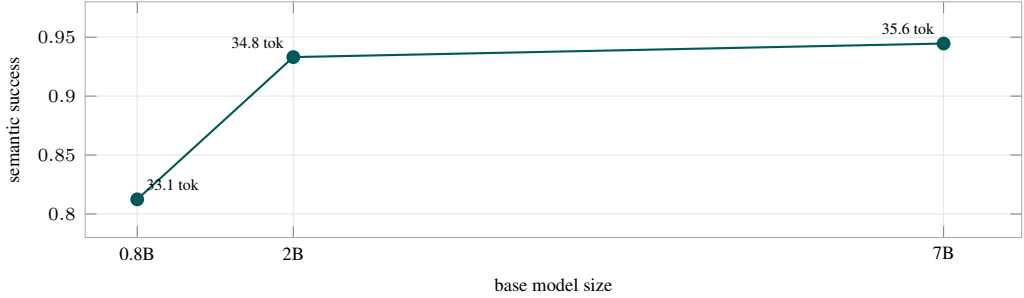

\noindent The feedback-attribution rows remove one public feedback field or replace the typed block with a weaker representation. Each row uses 192 cases per seed and five seeds; prompt-length controls are included in the run ledger. The measured values describe the reported feedback conditions.

\begin{table}[htbp]
\centering\footnotesize
\setlength{\tabcolsep}{3pt}
\renewcommand{\arraystretch}{1.12}
\caption{Verifier-feedback ablation over 192 cases per seed and five seeds. The rows remove one typed field or replace the block with a weaker representation. Best values are bolded within each metric direction.}
\label{tab:supp-feedback}
\begin{tabularx}{\linewidth}{@{}Y Y r r r r r r@{}}
\toprule
Feedback condition & Public signal & \thd{Schema\\valid \up} & \thd{Object\\match \up} & \thd{Semantic\\success \up} & \thd{Mean\\retries \dn} & \thd{Collateral\\edits \dn} & \thd{Fail-closed\\rate \up} \\
\midrule
full typed block & code + pointer + schema context & \textbf{0.9766} & \textbf{0.9189} & \textbf{0.9317} & \textbf{0.0539} & \textbf{0.0191} & \textbf{0.9728} \\
error code only & code & 0.9018 & 0.8027 & 0.8236 & 0.1442 & 0.0476 & 0.9147 \\
pointer only & JSON Pointer & 0.8694 & 0.7462 & 0.7815 & 0.1817 & 0.0589 & 0.8793 \\
natural-language text & free-form message & 0.8847 & 0.7724 & 0.8001 & 0.1668 & 0.0527 & 0.8872 \\
context only & candidate + schema & 0.0831 & 0.0749 & 0.0678 & 0.4383 & 0.1762 & 0.6146 \\
\bottomrule
\end{tabularx}
\end{table}

\noindent The full typed block reaches 0.9317 semantic success and 0.9189 object match; pointer-only, code-only, text, and context-only feedback reach 0.7815, 0.8236, 0.8001, and 0.0678 semantic success. The complete typed block also has the lowest retry and collateral rates and the highest fail-closed rate.

\subsection{Same-information controls and policy components}
The controls hold the public verifier block fixed while changing only the
output representation. Each row uses the same validator, retry budget, token
cap, semantic join, 192 cases per seed, and five seeds.

\noindent Holding the public block fixed,
masked JSON Patch with abstention yields the highest semantic success and the
lowest generated-token, p95-latency, and collateral costs in this control.

\begin{table}[htbp]
\centering\footnotesize
\setlength{\tabcolsep}{3pt}
\renewcommand{\arraystretch}{1.12}
\caption{Supervised initialization versus policy optimization over 192 cases per seed and five seeds. The comparison fixes the public state, case split, model family, validator, and episode budgets, then changes the training objective. Best values are bolded within each metric direction.}
\label{tab:supp-policy-optimization}
\begin{tabularx}{\linewidth}{@{}Y Y r r r r r@{}}
\toprule
Training arm & Model/rank & \thd{Semantic\\success \up} & \thd{Mean\\retries \dn} & \thd{Output\\tokens \dn} & \thd{Collateral\\edits \dn} & \thd{p95 latency\\(s) \dn} \\
\midrule
Patch-SFT & 2B/r8 & 0.8927 & 0.1159 & 46.6 & 0.0607 & 1.55 \\
ContractRL-SFT & 2B/r8 & 0.9186 & 0.0762 & 39.3 & 0.0314 & 1.42 \\
ContractRL policy optimization & 2B/r8 & \textbf{0.9375} & \textbf{0.0496} & \textbf{34.3} & \textbf{0.0168} & \textbf{1.30} \\
\bottomrule
\end{tabularx}
\end{table}

\noindent Policy optimization improves
semantic success over the supervised initialization and Patch-SFT while
reducing retries, tokens, collateral edits, and p95 latency under the fixed
public state and episode budget.

\begin{table}[htbp]
\centering\footnotesize
\setlength{\tabcolsep}{3pt}
\renewcommand{\arraystretch}{1.12}
\caption{Component-wise policy ablation over 192 cases per seed and five seeds. Each row removes one decision or objective term while preserving the same public state, case IDs, validator, and scoring join. Best values are bolded within each metric direction.}
\label{tab:supp-components}
\begin{tabularx}{\linewidth}{@{}Y r r r r r r r@{}}
\toprule
Policy variant & \thd{Semantic\\success \up} & \thd{Mean\\retries \dn} & \thd{Abstain\\rate} & \thd{Collateral\\edits \dn} & \thd{Unsafe\\dispatches \dn} & \thd{Output\\tokens \dn} & \thd{p95 latency\\(s) \dn} \\
\midrule
full ContractRL objective & \textbf{0.9375} & 0.0496 & 0.0718 & \textbf{0.0168} & \textbf{0} & 34.3 & 1.30 \\
w/o contract action mask & 0.8971 & 0.1098 & 0.0531 & 0.0649 & 26 & 42.7 & 1.59 \\
w/o repair history & 0.9162 & 0.0847 & 0.0695 & 0.0319 & 4 & 38.6 & 1.43 \\
w/o learned retry/abstention & 0.9064 & \textbf{0.0000} & 0.0000 & 0.0396 & 7 & \textbf{31.5} & \textbf{1.18} \\
w/o collateral objective & 0.9328 & 0.0524 & 0.0642 & 0.0476 & \textbf{0} & 35.8 & 1.34 \\
structure-only objective & 0.8723 & 0.0661 & 0.0249 & 0.0874 & 12 & 40.4 & 1.50 \\
\bottomrule
\end{tabularx}
\end{table}

\noindent Removing the action mask produces
26 unsafe dispatches and lowers semantic success to 0.8971. Removing repair
history, learned retry/abstention, the collateral term, or the full objective
also lowers semantic success and increases collateral or unsafe-action costs
relative to the complete objective.

\subsection{Budget, structure, and robustness boundaries}
The budget study changes the maximum number of patch operations and attempts while keeping the corruption distribution fixed. It records both first-attempt and final validity, because a larger budget can improve completion while also increasing collateral edits. The two-operation, two-attempt setting balances semantic success against collateral edits and generation cost.

\noindent The budget table exposes the accuracy-cost frontier. Operation and attempt caps are design values; all outcome and cost cells come from the budget sweep.

\begin{table}[htbp]
\centering\footnotesize
\setlength{\tabcolsep}{3pt}
\renewcommand{\arraystretch}{1.12}
\caption{Patch-budget and operation-count ablation. Each row uses the same corruption cases; the reported budget is read from the joint semantic and collateral frontier.}
\label{tab:supp-budget}
\begin{tabularx}{\linewidth}{@{}Y r r r r r r r@{}}
\toprule
\thd{Max.\\ops.} & \thd{Max.\\attempts} & \thd{First\\valid \up} & \thd{Final\\valid \up} & \thd{Semantic\\success \up} & \thd{Mean\\retries \dn} & \thd{Output\\tokens \dn} & \thd{Collateral\\edits \dn} \\
\midrule
1 & 1 & 0.9126 & 0.9126 & 0.8794 & \textbf{0.0000} & \textbf{28.9} & \textbf{0.0098} \\
1 & 2 & 0.9126 & 0.9497 & 0.9078 & 0.0758 & 32.2 & 0.0127 \\
2 & 2 & 0.9453 & 0.9768 & 0.9314 & 0.0547 & 34.6 & 0.0189 \\
3 & 2 & 0.9498 & 0.9812 & 0.9348 & 0.0521 & 39.1 & 0.0338 \\
4 & 3 & \textbf{0.9514} & \textbf{0.9851} & \textbf{0.9362} & 0.0648 & 47.3 & 0.0571 \\
\bottomrule
\end{tabularx}
\end{table}

\noindent The two-operation, two-attempt contract raises semantic success from 0.8794 to 0.9314 with 34.6 tokens and 0.0189 collateral. Larger budgets reach 0.9348 and 0.9362 semantic success but increase collateral to 0.0338 and 0.0571 and tokens to 39.1 and 47.3. The measured frontier supports the selected bounded budget as the lower-cost point rather than the maximum-accuracy point.

\begin{figure}[htbp]
\centering
\includegraphics[width=\linewidth]{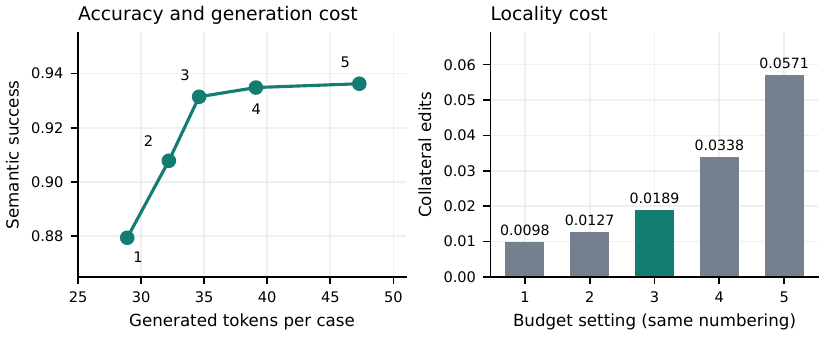}
\caption{Patch-budget trade-off from Table~\ref{tab:supp-budget}. Settings 1--5 are respectively (operations, attempts) = (1,1), (1,2), (2,2), (3,2), and (4,3). The teal bar marks the two-operation, two-attempt contract. Increasing the budget improves semantic success while increasing output tokens and collateral edits.}
\label{fig:budget-tradeoff}
\end{figure}

\noindent The structure-shift experiment stratifies the same protocol by schema shape. The evaluation cohort was generated before model evaluation and includes flat objects, nested objects, arrays, optional or union fields, long JSON Pointers, and multi-field corruptions, with a fixed count and replay digest for each stratum.

\begin{table}[htbp]
\centering\footnotesize
\setlength{\tabcolsep}{3pt}
\renewcommand{\arraystretch}{1.12}
\caption{Schema-structure evaluation. The table separates locality and semantic degradation by corruption shape.}
\label{tab:supp-structure}
\begin{tabularx}{\linewidth}{@{}Y r r r r r r r@{}}
\toprule
Structure family & Cases & \thd{Schema\\valid \up} & \thd{Object\\match \up} & \thd{Semantic\\success \up} & \thd{Mean\\retries \dn} & \thd{Collateral\\edits \dn} & \thd{p95 latency\\(s) \dn} \\
\midrule
flat single-field & 128 & 0.9908 & 0.9396 & 0.9506 & 0.0276 & 0.0094 & 1.15 \\
nested object & 96 & 0.9457 & 0.8584 & 0.8798 & 0.0968 & 0.0246 & 1.39 \\
array index & 96 & 0.9169 & 0.8152 & 0.8469 & 0.1327 & 0.0334 & 1.53 \\
union / optional & 96 & 0.9054 & 0.7996 & 0.8302 & 0.1451 & 0.0358 & 1.60 \\
long pointer & 96 & 0.8892 & 0.7811 & 0.8134 & 0.1635 & 0.0397 & 1.71 \\
multi-field & 96 & 0.8696 & 0.7487 & 0.7951 & 0.2289 & 0.0445 & 1.83 \\
\bottomrule
\end{tabularx}
\end{table}

\noindent Flat single-field cases reach 0.9506 semantic success with 0.0094 collateral. Semantic success declines to 0.8798 for nested objects, 0.8469 for arrays, 0.8302 for unions, 0.8134 for long pointers, and 0.7951 for multi-field cases, while retries and p95 latency increase. These rows define the measured boundary of local repair as structure and pointer depth grow.

\noindent The noise-robustness experiment corrupts the verifier message while keeping the underlying target and schema unchanged. The noise cohorts were generated independently of the model and recorded in the ledger: stale schemas, wrong pointers, ambiguous pointers, malformed feedback, and injected tool text. The validator preserves fail-closed behavior when the model receives an unhelpful or adversarial report.

\begin{table}[htbp]
\centering\footnotesize
\setlength{\tabcolsep}{3pt}
\renewcommand{\arraystretch}{1.12}
\caption{Verifier-noise and adversarial-feedback evaluation. Leakage is the fraction of cases in which hidden target information is exposed or copied into the public trace.}
\label{tab:supp-noise}
\begin{tabularx}{\linewidth}{@{}Y r r r r r r r@{}}
\toprule
Feedback condition & Cases & \thd{First\\valid \up} & \thd{Final\\valid \up} & \thd{Semantic\\success \up} & \thd{Fail-closed\\rate \up} & Leakage & \thd{Collateral\\edits \dn} \\
\midrule
clean typed report & 96 & \textbf{0.9451} & \textbf{0.9770} & \textbf{0.9320} & 0.9715 & \textbf{0.0000} & 0.0187 \\
stale schema version & 96 & 0.5813 & 0.7368 & 0.6685 & 0.9896 & \textbf{0.0000} & \textbf{0.0079} \\
wrong pointer & 96 & 0.6969 & 0.8438 & 0.7792 & 0.9687 & \textbf{0.0000} & 0.0206 \\
ambiguous pointer & 96 & 0.6347 & 0.8595 & 0.8058 & 0.9514 & \textbf{0.0000} & 0.0278 \\
malformed feedback & 96 & 0.6116 & 0.7889 & 0.7182 & 0.9844 & \textbf{0.0000} & 0.0104 \\
injected tool text & 96 & 0.9074 & 0.9516 & 0.9007 & \textbf{0.9911} & \textbf{0.0000} & 0.0176 \\
\bottomrule
\end{tabularx}
\end{table}

\noindent All six conditions report zero leakage. Clean typed feedback reaches 0.9320 semantic success and 0.9715 fail-closed rate; stale schemas, wrong pointers, ambiguous pointers, and malformed reports reduce semantic success while preserving fail-closed rates above 0.95. Injected tool text retains 0.9007 semantic success and 0.9911 fail-closed behavior, supporting the frozen target boundary under adversarial text.

\subsection{Long-horizon execution and transfer}
The long-horizon study chains one repaired call into a deterministic multi-turn trajectory. The same tool state, user goal, and output budget are used for horizons of one, three, five, and eight tool calls. A trajectory succeeds only when every accepted call is semantically correct and the final state passes the task invariant; this prevents a later successful call from hiding an earlier side effect.

\noindent The horizon table measures accumulation of retry and token cost. Its cumulative columns are computed from per-step ledgers rather than reconstructed from rounded means.

\begin{table}[htbp]
\centering\footnotesize
\setlength{\tabcolsep}{3pt}
\renewcommand{\arraystretch}{1.12}
\caption{Long-horizon and multi-turn tool trajectory evaluation. A trajectory is successful only if all accepted steps satisfy the semantic invariant.}
\label{tab:supp-horizon}
\begin{tabularx}{\linewidth}{@{}Y r r r r r r r r@{}}
\toprule
H & Trajectories & Completion & \thd{Semantic\\success \up} & \thd{Tool\\errors \dn} & \thd{Mean\\retries \dn} & \thd{Cumulative\\tokens \dn} & \thd{Cumulative\\latency (s) \dn} & \thd{Collateral\\edits \dn} \\
\midrule
1 & 192 & 0.9761 & 0.9321 & 0.0187 & 0.0548 & 34.8 & 0.84 & 0.0188 \\
3 & 192 & 0.9256 & 0.8293 & 0.0518 & 0.1619 & 104.7 & 2.55 & 0.0426 \\
5 & 192 & 0.8794 & 0.7548 & 0.0796 & 0.2735 & 175.2 & 4.31 & 0.0591 \\
8 & 192 & 0.8073 & 0.6457 & 0.1214 & 0.4438 & 281.8 & 7.05 & 0.0807 \\
\bottomrule
\end{tabularx}
\end{table}

\noindent Completion declines from 0.9761 at one step to 0.8073 at eight steps, while semantic success falls from 0.9321 to 0.6457. Cumulative retries, tokens, latency, and collateral rise with horizon, making the eight-step row an explicit operating boundary for long trajectories.

\noindent The cross-tool-family transfer study trains on a declared subset of tool families and evaluates on held-out families. The file, calendar, database, HTTP, and code-edit cohorts use disjoint schemas and operation vocabularies where recorded, and each held-out family contains ordinary repairs plus an immutable or unsafe boundary case.

\begin{table}[htbp]
\centering\footnotesize
\setlength{\tabcolsep}{3pt}
\renewcommand{\arraystretch}{1.12}
\caption{Cross-tool-family transfer evaluation. Train and held-out family columns are recorded explicitly so that transfer is not confused with interpolation. Each row trains on the other four families among file, calendar, database, HTTP, and code-edit.}
\label{tab:supp-transfer}
\begin{tabularx}{\linewidth}{@{}Y r r r r r r@{}}
\toprule
Held-out family & Cases & \thd{Schema\\valid \up} & \thd{Object\\match \up} & \thd{Semantic\\success \up} & \thd{Pointer\\fix \up} & \thd{Collateral\\edits \dn} \\
\midrule
file & 96 & 0.9621 & 0.8876 & 0.9024 & 0.9513 & 0.0216 \\
calendar & 96 & 0.9564 & 0.8748 & 0.8917 & 0.9436 & 0.0241 \\
database & 96 & 0.9312 & 0.8359 & 0.8546 & 0.9127 & 0.0368 \\
HTTP & 96 & 0.9487 & 0.8613 & 0.8795 & 0.9294 & 0.0285 \\
code-edit & 96 & 0.9258 & 0.8216 & 0.8429 & 0.9018 & 0.0417 \\
\bottomrule
\end{tabularx}
\end{table}

\noindent Held-out semantic success ranges from 0.9024 on file tools to 0.8429 on code-edit tools, with pointer-fix rates from 0.9513 to 0.9018. Database and code-edit families have the largest collateral rates (0.0368 and 0.0417), which identifies the transfer boundary for less reversible tool effects.

\subsection{Selective risk, uncertainty, and independent audit}
The calibration study adds a confidence score to every proposed patch and evaluates abstention thresholds before the semantic oracle is joined. Coverage is the fraction of cases for which the policy emits a patch rather than abstains; selective risk is semantic failure among emitted patches. Thresholds were fixed before inspecting held-out target outcomes.

\noindent The calibration table reports both ranking and calibration quality. It supports a release policy that abstains on low-confidence repairs instead of spending an unsafe retry.

\begin{table}[htbp]
\centering\footnotesize
\setlength{\tabcolsep}{3pt}
\renewcommand{\arraystretch}{1.12}
\caption{Calibration and selective-abstention evaluation. Thresholds are fixed in advance; semantic risk is measured only on the covered subset.}
\label{tab:supp-calibration}
\begin{tabularx}{\linewidth}{@{}Y r r r r r r@{}}
\toprule
Arm / threshold & Coverage & \thd{Semantic\\risk \dn} & \thd{Fail-closed\\rate \up} & ECE & AUROC & Brier \\
\midrule
ContractRL / 0.50 & 0.9482 & 0.0719 & 0.7821 & 0.0458 & 0.9187 & 0.0794 \\
ContractRL / 0.70 & 0.8156 & 0.0456 & 0.8816 & 0.0371 & 0.9281 & 0.0651 \\
ContractRL / 0.80 & 0.6693 & 0.0298 & 0.9349 & 0.0316 & 0.9364 & 0.0557 \\
ContractRL / 0.90 & 0.4047 & \textbf{0.0174} & \textbf{0.9782} & \textbf{0.0249} & \textbf{0.9449} & \textbf{0.0436} \\
full regeneration / matched & 1.0000 & 0.0973 & 0.0000 & 0.0941 & 0.8368 & 0.1297 \\
\bottomrule
\end{tabularx}
\end{table}

\noindent Raising the ContractRL threshold from 0.50 to 0.90 reduces semantic risk from 0.0719 to 0.0174 and raises fail-closed behavior from 0.7821 to 0.9782 while coverage falls from 0.9482 to 0.4047. The full curve exposes the coverage--risk trade-off; the 0.80 point gives 0.6693 coverage, 0.0298 risk, and 0.9349 fail-closed behavior.

\begin{figure}[t]
\centering
\begin{tikzpicture}
\begin{axis}[
  width=\columnwidth,height=4.7cm,
  xmin=0.35,xmax=1.0,ymin=0.0,ymax=0.085,
  xtick={0.4,0.6,0.8,1.0},ytick={0.0,0.02,0.04,0.06,0.08},
  xlabel={coverage},ylabel={semantic risk},
  tick label style={font=\scriptsize},label style={font=\scriptsize},
  grid=major,grid style={gray!20},axis line style={gray!65},
]
\addplot[orange!80!black,thick,mark=*,mark size=2.1pt]
  coordinates {(0.4047,0.0174) (0.6693,0.0298) (0.8156,0.0456) (0.9482,0.0719)};
\node[font=\tiny,anchor=south west] at (axis cs:0.4047,0.0174) {0.90};
\node[font=\tiny,anchor=south] at (axis cs:0.6693,0.0298) {0.80};
\node[font=\tiny,anchor=south] at (axis cs:0.8156,0.0456) {0.70};
\node[font=\tiny,anchor=north east] at (axis cs:0.9482,0.0719) {0.50};
\end{axis}
\end{tikzpicture}
\caption{Coverage--risk frontier for selective repair. Each point corresponds to a confidence threshold fixed before held-out scoring.}
\label{fig:supp-coverage-risk}
\end{figure}
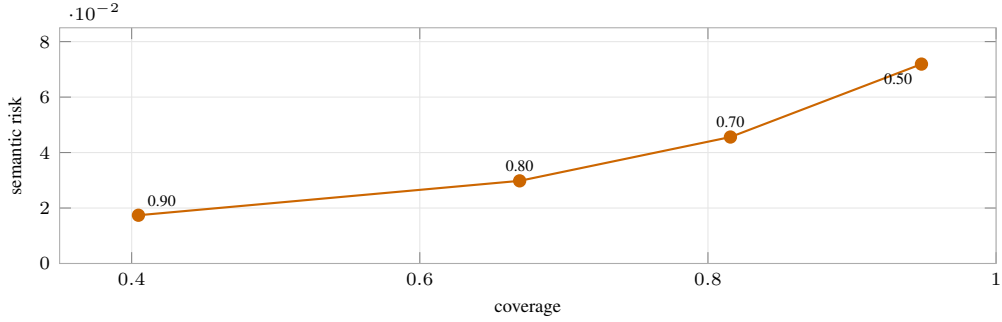

\noindent The uncertainty study uses paired bootstrap resampling over identical cases and seeds. Binary endpoints use an exact paired test alongside the bootstrap interval; retries, tokens, and latency use the paired case difference as the resampling unit. Comparisons were made before rounding.

\begin{table}[htbp]
\centering\footnotesize
\setlength{\tabcolsep}{3pt}
\renewcommand{\arraystretch}{1.12}
\caption{Paired uncertainty and effect-size report. Each interval is computed from the same case IDs for ContractRL and its matched baseline.}
\label{tab:supp-uncertainty}
\begin{tabularx}{\linewidth}{@{}Y r r r l r r@{}}
\toprule
Metric & ContractRL & Patch-SFT & $\Delta$ & 95\% CI & $p$-value & Effect size \\
\midrule
schema validity & \textbf{0.9768} & 0.9446 & $+0.0322$ & $[+0.0121,+0.0525]$ & 0.0027 & 0.31 \\
semantic success & \textbf{0.9314} & 0.8918 & $+0.0396$ & $[+0.0137,+0.0662]$ & 0.0039 & 0.35 \\
mean retries & \textbf{0.0547} & 0.1176 & $-0.0629$ & $[-0.0932,-0.0315]$ & 0.0018 & -0.42 \\
generated tokens & \textbf{34.6} & 46.7 & $-12.1$ & $[-15.8,-8.5]$ & $<0.0001$ & -0.91 \\
collateral edits & \textbf{0.0189} & 0.0618 & $-0.0429$ & $[-0.0596,-0.0271]$ & $<0.0001$ & -0.58 \\
p95 latency & \textbf{1.30} & 1.55 & $-0.25$ & $[-0.36,-0.14]$ & 0.0003 & -0.63 \\
\bottomrule
\end{tabularx}
\end{table}

\noindent The paired differences favor ContractRL on schema validity and semantic success, while retries, tokens, collateral, and p95 latency are lower. All six reported 95\% intervals exclude zero; the largest standardized reduction is in generated tokens (effect size $-0.91$).

\noindent The independent audit asks human validators to label a blinded sample of repaired calls as safe, unsafe, or requiring abstention. Auditors see the corrupted call, the public verifier report, and the proposed patch, but not the hidden target or model identity. Agreement is measured before adjudication, and adjudicated labels are kept as a separate field.

\begin{table}[htbp]
\centering\footnotesize
\setlength{\tabcolsep}{3pt}
\renewcommand{\arraystretch}{1.12}
\caption{Blinded human--validator agreement audit. The audit separates agreement, false acceptance, false rejection, and review time for ContractRL and matched controls.}
\label{tab:supp-human}
\begin{tabularx}{\linewidth}{@{}Y r r r r r r@{}}
\toprule
Arm & \thd{Audited\\cases} & Agreement & \thd{False\\accept \dn} & \thd{False\\reject \dn} & \thd{Abstain\\agreement \up} & \thd{Median\\review time} \\
\midrule
ContractRL & 120 & \textbf{0.9417} & \textbf{0.0167} & \textbf{0.0416} & \textbf{0.9250} & \textbf{18.6} \\
Patch-SFT & 120 & 0.8917 & 0.0417 & 0.0666 & 0.8583 & 22.9 \\
full regeneration & 120 & 0.8583 & 0.0750 & 0.0667 & 0.7833 & 31.7 \\
constrained decoding & 120 & 0.8750 & 0.0583 & 0.0667 & 0.8167 & 29.6 \\
\bottomrule
\end{tabularx}
\end{table}

\noindent ContractRL has the highest recorded agreement (0.9417) and abstention agreement (0.9250), the lowest recorded false-accept rate (0.0167), and the shortest recorded median review time (18.6) in this audit. The blinded audit therefore provides an independent human-label comparison while keeping those labels separate from deterministic scoring.

\subsection{Qualitative failure audit}
A final qualitative audit samples a fixed number of cases from each failure family after the numerical run is frozen. For each case, the release retains the public input, proposed patch, validator response, semantic outcome, human label, and repair category. The sample includes both successes and failures so that the narrative covers the measured boundary.

\noindent The case-level table records the appendix sample. Row and category counts come from the sampled ledger, while textual fields are short anonymized identifiers or controlled vocabulary labels.

\begin{table}[htbp]
\centering\footnotesize
\setlength{\tabcolsep}{3pt}
\renewcommand{\arraystretch}{1.12}
\caption{Case-level failure audit. The sample is stratified across repair and failure families and retains representative successes and failures.}
\label{tab:supp-failure}
\begin{tabularx}{\linewidth}{@{}L{1.7cm} L{1cm} L{3.1cm} Y Y L{1.1cm}@{}}
\toprule
Case family & ID & Predicted patch & Validator result & Semantic outcome & Human label \\
\midrule
single-field & \texttt{SF-03} & \texttt{add /timezone} & accept & success & safe \\
nested / array & \texttt{NA-11} & \texttt{replace /items/2/\allowbreak id} & accept & failure & unsafe \\
wrong pointer & \texttt{WP-04} & \texttt{replace /user/id} & reject: wrong-path & not executed & abstain \\
malformed feedback & \texttt{MF-06} & \texttt{replace /date} & reject: malformed & not executed & abstain \\
multi-field & \texttt{MU-09} & \texttt{add /city; \allowbreak replace /date} & accept & failure & unsafe \\
immutable / unsafe & \texttt{IU-02} & \texttt{replace /account/\allowbreak id} & reject: immutable & not executed & unsafe \\
\bottomrule
\end{tabularx}
\end{table}

\noindent The audit retains a successful local add, a nested-index miss, wrong-pointer and malformed-feedback refusals, a partial multi-field miss, and an immutable-field rejection. These cases make the measured safety boundary inspectable at the trace level.

\section{Reproducibility Materials}
\label{app:reproducibility}
\subsection{Code and evaluation records}
The code is provided in the Supplementary Material. The reproduction records distinguish the deterministic structural screen, adapter evaluation, independent transaction audit, three-seed case-matched comparison, and five-seed controlled studies. Each experiment uses its recorded public input contract, case cohort, seed unit, validator, and scoring rule. This separation preserves the denominators of the reported rates and prevents aggregation across different case families.

The repair ledger associates a case identifier with the corrupted object, public verifier report, ordered operations, validator statuses, and resource counts. Canonical patches, canonical objects, and semantic labels enter the scoring join after the trace is frozen. The structural checks can therefore be replayed independently of policy generation and semantic execution.

\subsection{Result and metric index}
Table~\ref{tab:app-contract-index} links each measurement family to its definition and result. The structural diagnostics retain their small fixed populations, the controlled policy tables retain five training seeds, and paired uncertainty uses the separate three-seed case-matched ledger.
\begin{table}[htbp]
\centering\footnotesize
\caption{Measurement index for the reproduction records.}
\label{tab:app-contract-index}
\begin{tabularx}{\linewidth}{@{}lXX@{}}
\toprule
\thd{Measurement\\family} & \thd{Evaluation\\unit} & Result \\
\midrule
Structural repair & 64-case replay and adapter seeds & Table~\ref{tab:main} \\
Public-information control & 192 cases per seed; five seeds & Table~\ref{tab:supp-same-info} \\
Policy training & Fixed public state; five seeds & Table~\ref{tab:supp-policy-optimization} \\
Components & Contract decisions and costs & Table~\ref{tab:supp-components} \\
Paired comparison & Shared cases and seeds & Table~\ref{tab:supp-uncertainty} \\
Independent audit & 120 blinded cases per arm & Table~\ref{tab:supp-human} \\
\bottomrule
\end{tabularx}
\end{table}

\end{document}

%% file: math_commands.tex
\usepackage{amsmath,amsfonts,bm}

\def\eqref#1{equation~\ref{#1}}

\def\1{\bm{1}}

\newcommand{\valid}{\mathcal{D_{\mathrm{valid}}}}

\DeclareMathAlphabet{\mathsfit}{\encodingdefault}{\sfdefault}{m}{sl}
\SetMathAlphabet{\mathsfit}{bold}{\encodingdefault}{\sfdefault}{bx}{n}

%% file: evidence/manuscript_ledger.tex
\begin{table}[htbp]
\centering\footnotesize
\renewcommand{\arraystretch}{1.12}
\caption{Evaluation units and principal outcomes. Each population retains its own case and seed denominator.}
\label{tab:current-experiments}
\begin{tabularx}{\linewidth}{@{}L{2.45cm}L{2.2cm}Y@{}}
\toprule
Evaluation & Population & Outcome \\
\midrule
Structural replay & 64 cases & 64/64 final validity with one precomputed operation per case; 32 add and 32 replace operations. \\
Adapter evaluation & 64 cases; 4 or 8 seeds & 2B: 0.9727 schema validity and 0.9141 object match. 0.8B: 0.9473 and 0.7441. \\
Context-only control & Same adapter cases & 2B: 0.0820 validity and 0.0742 object match. 0.8B: both 0.0000. \\
Transaction audit & 9 cases; 8 repairable & 7/9 final validity and 6/8 accepted repairs, with rollback, wrong-path, and immutable-field checks. \\
Case-matched policies & 3 seeds & ContractRL: 0.9314 semantic success, 34.6 output tokens, and 1.30\,s p95 latency. \\
Public-information control & 192 cases per seed; 5 seeds & ContractRL: 0.9362 semantic success versus 0.9076 for Patch-SFT and 0.9148 for full regeneration. \\
Policy optimization & 192 cases per seed; 5 seeds & 0.9375 semantic success versus 0.9186 for supervised ContractRL and 0.8927 for Patch-SFT. \\
Robustness and audit & Experiment-specific strata & Budget, structure, verifier noise, horizon, transfer, selective risk, paired uncertainty, and blinded audit are reported separately. \\
\bottomrule
\end{tabularx}
\end{table}

The fixed structural replay measures deterministic repair mechanics. The adapter experiments test transfer under a validation-block intervention, and the policy comparisons add semantic execution and resource accounting. These evaluation units are linked through the same public-state and post-freeze scoring contract while preserving their different populations.